\def\year{2020}\relax
\documentclass[letterpaper]{article} % DO NOT CHANGE THIS
\usepackage{aaai20}  % DO NOT CHANGE THIS
\usepackage{times}  % DO NOT CHANGE THIS
\usepackage{helvet} % DO NOT CHANGE THIS
\usepackage{courier}  % DO NOT CHANGE THIS
\usepackage[hyphens]{url}  % DO NOT CHANGE THIS
\usepackage{graphicx} % DO NOT CHANGE THIS
\usepackage{graphicx}  % DO NOT CHANGE THIS
\newcommand{\RNum}[1]{\uppercase\expandafter{\romannumeral #1\relax}}
\usepackage{amsmath}
\usepackage{amsfonts}
\usepackage{blindtext}
\usepackage{algorithm}
\usepackage{algpseudocode}
\usepackage{subfigure}
\usepackage{enumitem}
\usepackage{epstopdf}
\usepackage{multirow}
\usepackage{booktabs}

\title{Multi-Agent Actor-Critic\\ with Hierarchical Graph Attention Network}
\author{\Large \textbf{Heechang Ryu, Hayong Shin, Jinkyoo Park\thanks{Corresponding author}}\\ % All authors must be in the same font size and format. Use \Large and \textbf to achieve this result when breaking a line
Industrial \& Systems Engineering, KAIST, Republic of Korea\\ %If you have multiple authors and multiple affiliations
\{rhc93, hyshin, jinkyoo.park\}@kaist.ac.kr
}
 
\begin{document}

\maketitle

\begin{abstract}
Most previous studies on multi-agent reinforcement learning focus on deriving decentralized and cooperative policies to maximize a common reward and rarely consider the transferability of trained policies to new tasks. This prevents such policies from being applied to more complex multi-agent tasks. To resolve these limitations, we propose a model that conducts both representation learning for multiple agents using hierarchical graph attention network and policy learning using multi-agent actor-critic. The hierarchical graph attention network is specially designed to model the hierarchical relationships among multiple agents that either cooperate or compete with each other to derive more advanced strategic policies. Two attention networks, the inter-agent and inter-group attention layers, are used to effectively model individual and group level interactions, respectively. The two attention networks have been proven to facilitate the transfer of learned policies to new tasks with different agent compositions and allow one to interpret the learned strategies. Empirically, we demonstrate that the proposed model outperforms existing methods in several mixed cooperative and competitive tasks.
\end{abstract}

\section{Introduction}
In nature, the battle for dominance rights or desirable territories between individuals is a typical phenomenon \cite{smith1973logic}. Occasionally, individuals in the same group cooperate to compete against enemy groups. They can gain stronger immunity against predators \cite{ugelvig2007social} or powerful forces to overpower preys \cite{powell2004combat}.
% ; this phenomenon prevails in human society \cite{puurtinen2008between}.
The cooperation and competition among agents are also important modeling paradigms in various engineering systems, such as smart grids \cite{dall2013distributed}, logistics \cite{ying2005multi,cao2013overview}, and distributed vehicles$/$robots  \cite{corke2005networked,fax2004information,matignon2012coordinated}. To control such complex systems composed of many interacting components, researchers have studied multi-agent reinforcement learning (MARL) for a long time.
% and is now attracting more and more attention.

Recently, reinforcement learning (RL) combined with deep neural networks has achieved human-level or higher performances in challenging games \cite{mnih2015human,silver2016mastering,silver2017mastering}. The advances in RL and deep learning have led great interests in MARL, hoping that it can resolve complex and large scale problems. The majority of MARL algorithms have focused on deriving decentralized policies for conducting a collaborative task. For the collection of actions individually determined by decentralized policies to be coordinated, it is important to impose consensus when deriving the decentralized policies. To achieve this, under the concept of centralized training and decentralized execution (CTDE), MARL learns centralized critics for multiple agents and derives decentralized actors using partial gradient from the centralized critics. Depending on the information available in the execution phase, CTDE approach can be further categorized into ``learning-for-consensus approach", where each agent determines decentralized action solely based on its local observation \cite{lowe2017multi,foerster2018counterfactual,zhang2018fully,iqbal2019actor}, and ``learning-to-communicate approach", where each agent employs communication schemes (i.e., interchanges messages) 
% to achieve consensus among agents
during the execution phase \cite{sukhbaatar2016learning,peng2017multiagent,jiang2018learning,IC3Net,das2019tarmac}.

Although a variety of MARL algorithms have been proposed, unresolved issues still exist in MARL when it is applied to realistic environments. First, in terms of the generality of the modeling framework, most models focus on deriving pure cooperation or competition among multiple agents by forcing all agents to seek a shared common reward rather than modeling the relationships among heterogeneous agents in a mixed cooperative-competitive task (issue in modeling flexibility). In addition, the models are limited for modeling a large number of agents owing to the curse of dimensionality in modeling centralized critics (issue in scalability). This is connected with a more fundamental limitation, in which the trained model cannot be transferred to different tasks with different numbers of agents having different goals (issue in transferability).

We herein propose a model, called \textbf{H}ierarchical graph \textbf{A}ttention-based \textbf{M}ulti-\textbf{A}gent actor-critic (HAMA), that conducts both representation learning for multi-agent system and policy learning using multi-agent actor-critic in end-to-end learning. HAMA employs a hierarchical graph neural network to effectively model the inter-agent relationships in each group of agents and inter-group relationships among groups. HAMA additionally employs inter-agent and inter-group attentions to adaptively extract the state-dependent relationships among multiple agents, which is proven to be effective for helping policies to adjust their high-level strategies (e.g., cooperate or compete). The combination of hierarchical graph neural networks with two distinct attention layers, which we refer to as a \textbf{H}ierarchical \textbf{G}raph \textbf{A}ttention ne\textbf{T}work (HGAT), effectively processes the local observation of each agent into a single embedding vector, an information-condensed and contextualized state representation for each individual agent. HAMA sequentially uses the embedding vector for each agent to compute the individual critic and actor for deriving decentralized policies. %Owing to the use of HGAT, HAMA's network is scalable to a large-scale environment with many agents, and the trained policy is transferable to new environments, and the derived policy is interpretable.

We empirically demonstrate that HAMA outperforms existing MARL algorithms in four different game scenarios. Furthermore, we demonstrate that the policies trained by HAMA in a small-scale game with a small number of agents can be applied directly to control a large number of agents in a new game. Finally, we demonstrate that inter-agent and inter-group attentions can be used to interpret the derived policy and decision-making process.

\section{Related Work}
We categorize existing MARL studies into two categories; learning-for-consensus approach and learning-to-communicate approach depending on how the consensus among multiple agents is derived.
\subsubsection{Learning-for-Consensus Approaches.} Learning-for-consensus approaches in MARL focus on deriving decentralized policies (actors) for agents, each of which maps a local observation for an agent to an individual action for it. To make such individually chosen actions be coordinated to conduct collaborative tasks, these approaches first construct a centralized critic for either a global reward or individual reward and use the centralized critic to derive the decentralized actor.
% \cite{lowe2017multi,foerster2018counterfactual,zhang2018fully,iqbal2019actor}.
MADDPG \cite{lowe2017multi} has extended DDPG \cite{lillicrap2015continuous} to multi-agent settings for mixed cooperative-competitive environments. COMA \cite{foerster2018counterfactual} constructs a centralized critic and computes an agent-specific advantage function to derive a decentralized actor. FDMARL \cite{zhang2018fully} has proposed a distributed learning approach for each agent to learn a global critic using its local reward and the transferred critic parameters from the networked neighboring agents. Because these models directly use the state or observation in constructing critic or actor networks, it is difficult to apply such models to a large-scale environment or transfer them to new environments. As a way to resolve the scalability issue, the concepts of graph neural network \cite{gori2005new,scarselli2009graph,battaglia2018relational} and attention network have been employed to effectively represent the global state and accordingly centralized critics.
% and action, and accordingly centralized critics. 
For example, MAAC \cite{iqbal2019actor} employs the attention network and graph neural network to model a centralized critic, from which decentralized actors are derived using soft actor-critic \cite{haarnoja2018soft}. In addition, DGN \cite{jiang2018graph} applies a graph convolutional network to model a centralized Q-function for each agent using a deep Q-network \cite{mnih2015human}.
%, from which individual action is selected in a decentralized manner.

% The majority of MARL algorithms focus on deriving decentralized policies for conducting a collaborative task. For the collection of actions individually chosen by decentralized policies to be coordinated, it is important to impose consensus when deriving the decentralized policies. One method to achieve the consensus is to construct a centralized critic for either a global reward or individual reward and use the centralized critic to derive the decentralized actor \cite{lowe2017multi,foerster2018counterfactual,zhang2018fully}. MADDPG \cite{lowe2017multi} has extended deep deterministic policy gradient (DDPG) \cite{lillicrap2015continuous} to multi-agent settings for mixed cooperative-competitive environments. COMA \cite{foerster2018counterfactual} constructs a centralized critic and computes an agent-specific advantage function to derive a decentralized actor. FDMARL \cite{zhang2018fully} proposes a distributed learning approach for each agent to learn (approximate) a global critic using its local reward and the transferred critic parameters from the networked neighboring agents. Because these models directly use the state or observation in constructing critic or actor networks, it is difficult to apply such models to a large-scale environment or transfer them to new environments.

\subsubsection{Learning-to-Communicate Approaches.} Another method to achieve consensus among decentralized policies in cooperative environments is using communication among agents. In this framework, each agent learns how to transmit messages to other agents and process the messages received from other agents to determine an individual action. During the centralized training, such message generating and processing procedures are learned to induce cooperation among agents. During the execution phase, agents exchange the messages to determine their actions. CommNet \cite{sukhbaatar2016learning} uses a large single neural network to process all the messages transmitted by all agents globally, and the processed message is used to guide all agents to cooperate. BiCNet \cite{peng2017multiagent} has proposed a communication channel in the form of bi-directional recurrent network to accommodate messages from any number of agents, thus resolving the scalability issue. To effectively specify the communication structure while considering the relative relationships among agents, ATOC \cite{jiang2018learning} has proposed a communication channel in a bi-directional LSTM with an attention layer. The attention layer in ATOC enables each agent to process the messages from other agents differently depending on their state-dependent importance. Similar to ATOC, IC3Net \cite{IC3Net} has been proposed to actively select and mask messages from other agents during communication by applying gating function in the message aggregation step. TarMAC \cite{das2019tarmac} has proposed a targeted communication protocol to determine whom to communicate with and what messages to transmit using attention networks. Such communication-based methods use attention networks to learn the communication structure$/$protocol effectively. Although the communication helps each agent to use extensive information in its individual decision making, this approach requires a separate communication channel and a well-established communication environment for message exchange. Furthermore, some communication-based methods can be applied to only cooperative tasks because it does not make sense to receive messages from competitive agents while playing a game.

\subsubsection{Novelties of HAMA.} As introduced, the graph neural network and attention network structures have been widely employed (1) to model a critic for scalable learning in learning-for-consensus approach, and (2) to model communication structure in learning-to-communicate approach. HAMA, our proposed model, employs the HGAT to embrace the merits of employing a graph representation in both learning-for-consensus and learning-to-communicate approaches. The proposed HGAT capturing enhanced relative inductive biases \cite{battaglia2018relational} enables HAMA to model both centralized critic and decentralized actor (1) that can be scalable and transferable and (2) that can effectively utilize the contextualized state representation. In particular, because it learns how to represent the partial observation through graph embedding, it is different from other communication approaches that utilize the messages processed by other agents. Therefore, our approach is considered to be a kind of learning-for-consensus approaches and can be applied for mixed cooperative and competitive environments easily.

\begin{figure*}[t]
  \centering
  \includegraphics[scale = 0.40]{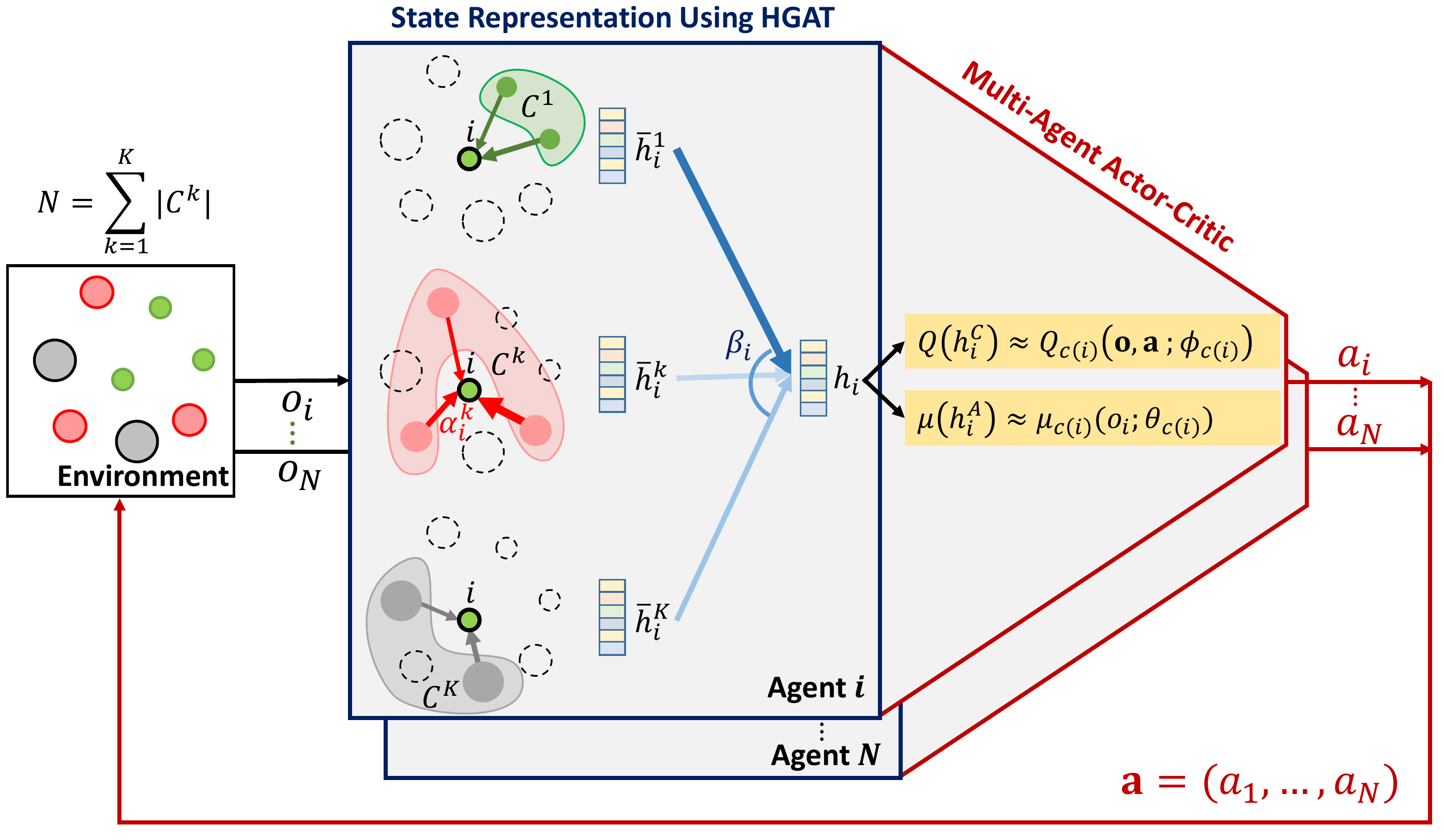}
  \caption{Overview of HAMA.}
  \label{fig:fig1}
\end{figure*}

\section{Background}
\subsubsection{Partially Observable Markov Game (POMG).} A POMG is an extension of partially observable Markov decision process to a game with multiple agents. A POMG for $\mathnormal{N}$ agents is defined as follows: $s\in\mathcal{S}$ denotes the global state of the game; ${o_i}\in\mathcal{O}_{i}$ denotes a local observation that agent $i$ can acquire; $a_i\in\mathcal{A}_{i}$ is an action for agent $i$. The reward for agent $i$ is computed as a function of state $s$ and joint action $\mathbf{a}$ as ${r}_{i}:\mathcal{S}\times\mathcal{A}_1\times\dots\times\mathcal{A}_N\mapsto\mathbb{R}$. The state evolves to the next state according to the state transition model $\mathcal{T}:\mathcal{S}\times\mathcal{A}_1\times\dots\times\mathcal{A}_N\mapsto\mathcal{S}$. The initial state is determined by the initial state distribution $\rho:\mathcal{S}\mapsto[0,1]$. The agent $i$ aims to maximize its discounted return $R_i=\sum_{t=0}^{T} \gamma^t r^t_i$, where $\gamma\in[0,1]$ is a discount factor.

\subsubsection{Multi-Agent Deep Deterministic Policy Gradient (MADDPG).} Deterministic policy gradient (DPG) \cite{silver2014deterministic} aims to directly derive a deterministic policy, $a=\mu(s;\theta)$, that maximizes the expected return $\mathcal{J(\theta)}=\mathbb{E}_{s\sim\rho^\mu,a\sim\mu_\theta}[R]\approx\mathbb{E}_{s\sim\rho^\mu,a\sim\mu_\theta}[Q^{\mu}(s,a;\phi)]$, where $Q^{\mu}(s,a;\phi) = \mathbb{E}_{s'}[r(s,a)+\gamma\mathbb{E}_{a'\sim\mu}[Q^\mu(s',a')]$. The parameter $\theta$ of $\mu(s;\theta)$ is subsequently optimized by the gradient of  $\mathcal{J(\theta)}$: $\nabla_\theta\mathcal{J}(\theta)=\mathbb{E}_{s\sim\mathcal{D}}[\nabla_\theta\mu(s;\theta){\nabla_a}Q^{\mu}(s,a;\phi)\arrowvert_{a=\mu(s;\theta)}]$. The $\mathcal{D}$ is an experience replay buffer that stores  $(s,a,r,{s'})$ samples. Deep deterministic policy gradient (DDPG), an actor-critic model based on DPG, uses deep neural networks to approximate the critic and actor of each agent.

MADDPG is a multi-agent extension of DDPG for deriving decentralized policies for the POMG. MADDPG comprises the individual $Q$-network and policy network for each agent. The $Q$-network for agent $i$ is learned by minimizing the loss: $\mathcal{L}(\phi_i)={\mathbb{E}_{\mathbf{o},\mathbf{a},r,{\mathbf{o}'}\sim \mathcal{D}}}[(Q_i^{\mu}(\mathbf{o},\mathbf{a};\phi_i)-y_i)^2],$ where $\mathbf{o}=(o_1,\dots,o_N)$ and $\mathbf{a}=(a_1,\dots,a_N)$ are, respectively, the observations and actions of all agents, and $y_i=r_i(s,\mathbf{a})+\gamma {Q_i^{\mu'}}({\mathbf{o}'},\mathbf{a}';{\phi_i}')\arrowvert_{a_j'=\mu'(o'_j);\theta')}$. The policy network $\mu_i(o_i;\theta_i)$ of agent $i$ is optimized using the gradient: $\nabla_{\theta_i}\mathcal{J}(\theta_i)
=\mathbb{E}_{\mathbf{o},\mathbf{a}\sim \mathcal{D}}[\nabla_{\theta_i}\mu_i(o_i;\theta_i)\nabla_{a_i}Q_i^{\mu}(\mathbf{o},\mathbf{a};\phi_i )\arrowvert_{a_i=\mu_i(o_i;\theta_i)}]$.

\subsubsection{Graph Attention Network (GAT).} The GAT \cite{velivckovic2017graph} is an effective model to process structured data that is represented as a graph. The GAT has proposed a way to compute the node-embedding vector of graph nodes by aggregating node embeddings $h_{j}$ from neighboring nodes $\{j\in \mathcal{N}_i\}$ that are connected to the target node $i$ as $h'_{i}=\sigma(\sum_{j\in \mathcal{N}_i} \alpha_{ij} \mathbf{W} h_j)$. The attention weight $\alpha_{ij}=\text{softmax}_{j}(e_{ij})$, where $e_{ij}=a(\mathbf{W}h_i,\mathbf{W}h_{j})$, quantifies the importance of node $j$ to node $i$ in computing node-embedding value $h'_i$.

\section{Methods}
HAMA comprises a representation learning framework for processing the state represented as a graph and a multi-agent actor-critic network for deriving decentralized policies for the agents. As shown in Figure \ref{fig:fig1}, HAMA represents the game state as a graph and computes for each agent the node-embedding vector that compactly summarizes each agent's status in relation with other groups of agents and environment. The computed node-embedding vector for each agent is subsequently used to compute the Q-value and action in an actor-critic framework. %(i.e., locations and features of all the agents)

\subsection{State Representation Using HGAT}
We propose HGAT, a network stacking multiple
% graph attention networks 
GATs hierarchically, that processes each agent's local observation into a high-dimensional node-embedding vector to represent the hierarchical inter-agent and inter-group relationships of each agent.

\subsubsection{Agent Clustering.} The first step in representation learning is to cluster all the agents into distinct groups $C^k$ using prior knowledge or data. For pure cooperative tasks, all the agents can be categorized into a single group. If the target task involves competition between two groups, we can cluster the agents into two groups. In addition, we can cluster into a group the agents that do not execute any actions but participate in the game (i.e., terrain components or obstacles). In this study, we assume that the agents can be easily clustered into $K$ groups using prior knowledge on the agents, which implies that HAMA utilizes enhanced relative inductive biases regarding the group relationships.

\subsubsection{Node-Embedding Using GAT in Each Cluster.} Agent $i$ has the local observation $o_i = \{s_j\arrowvert\ j \in V(i)\}$ where $s_j$ is the local state of agent $j$, and $V(i)$ specifies the visual range of agent $i$. The visual range can be specified depending on environment settings so that agent $i$ can observe the agents within a certain distance. Thus, our agent can observe nearby agents  as a partial observation. Agent $i$ computes the different node-embedding vectors $\bar{h}_i^k$ for different groups $k=1,...,K$ to summarize the individual relationships between agent $i$ and agents from different groups. To compute  $\bar{h}_i^k$,  agent $i$ first computes embedding $h^k_{ij}=f^k_M(s_i,s_j;w^k_M)$ between itself and agents in ${j\in C^k \cap V(i)}$ and computes the aggregated embedding $\bar{h}^k_i=\sum_{j \in {C^k \cap V(i)}} \alpha^k_{ij}h^k_{ij}$. The inter-agent attention weight $\alpha_{ij}^k$ quantifies the importance of the embedding $h^k_{ij}$ from agent $j$ to agent $i$. The inter-agent attention weight is computed as softmax  $\alpha^k_{i,\cdot}\propto\text{exp}(e^k_{i,\cdot})$ where  $e^k_{ij}=f^k_\alpha(s_i,s_j;w^k_\alpha)$. The attention can be extended to multiple attention heads \cite{vaswani2017attention}, but the current study employs only plain and classical attention networks.
% but the current study does not employ it because it does not improve the performance.
% The embedding function $f^k_M(s_i,s_j;w^k_M)$ and the attention function $f^k_\alpha(s_i,s_j;w^k_\alpha)$ are shared by the agents in the same group $k$, thus facilitating training and generalization.
It is noteworthy that agent $i$ computes embedding $h^k_{ij}$ by processing its own observation on other agents; therefore, the other agents are not required to send messages to agent $i$, unlike other learning-to-communicate approaches that require agents to exchange their hidden vectors. 

\subsubsection{Hierarchical State Representation Using Multi-Graph Attention.} This step aggregates the group-level node-embedding vectors $\bar{h}_i^1,...,\bar{h}_i^K$ of agent $i$ for the information-condensed and contextualized state representation of agent $i$ as $h_i=\sum_{k=1}^{K} {\beta^k_{i}\bar{h}^k_i}$ while considering the relationships between agent $i$ and the groups of other agents. The inter-group attention weight $\beta^k_{i}$ guides which group agent $i$ should focus more on to achieve its objective. For example, if $\beta^k_{i}$ is large for the same group which agent $i$ belongs to, it implies that agent $i$ focuses on cooperating with the agents in the same group. Otherwise, agent $i$ would focus more on competing with agents from different groups. The inter-group attention weight is computed as softmax $\beta_i=(\beta^1_{i},...,\beta^K_{i})\propto\text{exp}(q_{i})$ where $q_{i}=[q^1_{i},\dots,q^K_{i}]=f_\beta([\bar{h}^1_i,\dots,\bar{h}^K_i];w_\beta)$. The hierarchical state representation is particularly useful when considering mixed cooperative-competitive games where each agent or group possesses their own objectives, which will be empirically shown by various experimental results in this study. The embedding and attention functions in this study comprise a two-layered MLP with 256 units and ReLUs. 

\subsection{Multi-Agent Actor-Critic}
The proposed method uses the embedding vectors $h_i^C$ and $h_i^A$ of agent $i$ to compute, respectively, the individual Q-value $Q_{c(i)}(\mathbf{o},\mathbf{a})\approx Q_{c(i)}(h_i^C;\phi_{c(i)})$ and determine the action $a_i=\mu_{c(i)}(o_i)\approx \mu_{c(i)}(h_i^A;\theta_{c(i)})$, where $c(i)$ is the group to which agent $i$ belongs. Note that the embedding vectors $h_i^C$ and $h_i^A$ are computed separately using two different HGATs; computing $h_i^C$ requires a joint action $\mathbf{a}$ in the training phase under CTDE. Additionally, agents in the same group share the actor and critic networks for generalization.
% better training and generalization.

Compared to using raw observation as an input for the critic and actor network \cite{lowe2017multi}, using node-embedding vectors computed from HGAT as inputs offers the following advantages: (1) a node-embedding vector can be computed by considering the hierarchical relationships among agents, i.e., relative inductive biases, thus providing contextualized state representation; (2) it is scalable to a large number of agents as the dimension of a node-embedding vector does not change with the number of agents; and (3) HGAT enables the learned policy to be used in environments of any agent or group size, i.e., the property that transfer learning aims to achieve.

The training of HAMA is similar to that of MADDPG. The shared critic $Q_k$ for agent $i$ in group $k$ is trained to minimize the loss $\mathcal{L}$:
\begin{multline}
\mathcal{L}(\phi_k)={\mathbb{E}_{{\mathbf{o}},\mathbf{a},r_i,{\mathbf{o}'}\sim \mathcal{D}}}[(Q_k^{\mu}({\mathbf{o}},\mathbf{a};\phi_k)-y_i)^2],\\ y_i=r_i+{\gamma}Q_k^{\mu'}(\mathbf{o}',\mathbf{a}';\phi'_k)\arrowvert_{a_i'=\mu'(o_i';\theta')}
\end{multline}
where $Q^{\mu'}$ and $\mu'$ are, respectively, the target critic and actor networks for stable learning with delayed parameters
% , which is called  \emph{soft target} 
\cite{lillicrap2015continuous}. In CTDE framework, the joint observation and action are assumed to be available for training. The shared actor $\mu_k$ for agent $i$ in group $k$ is then trained using gradient ascent algorithm
% the gradient based optimization algorithm, e.g., $\theta_k\leftarrow \theta_k+\alpha\nabla_{\theta_k}\mathcal{J}(\theta_k)$, where $\alpha$ is learning rate ,
with the gradient computed as:
\begin{multline}
\nabla_{\theta_k}\mathcal{J}(\theta_k)
=\\
\mathbb{E}_{\mathbf{o},\mathbf{a}\sim \mathcal{D}}[\nabla_{\theta_k}\mu_k(o_i;\theta_k)\nabla_{a_i}Q_k^{\mu}(\mathbf{o},\mathbf{a};\phi_k )\arrowvert_{a_i=\mu_k(o_i;\theta_k)}] 
\end{multline}
where $a_i$ is the action of agent $i$ in $\mathbf{a}$. During the training, the joint observation $\mathbf{o}$ and joint action $\mathbf{a}$ are used, whereas, during the execution, only the learned policy
$\mu_k(o_i;\theta_k)\approx\mu_{c(i)}(h_i^A;\theta_{c(i)})$
is used with the embedding vector $h_i^A$ computed using only local observation $o_i$ of agent $i$.

\begin{figure}[t]
  \centering
  \includegraphics[scale = 0.45]{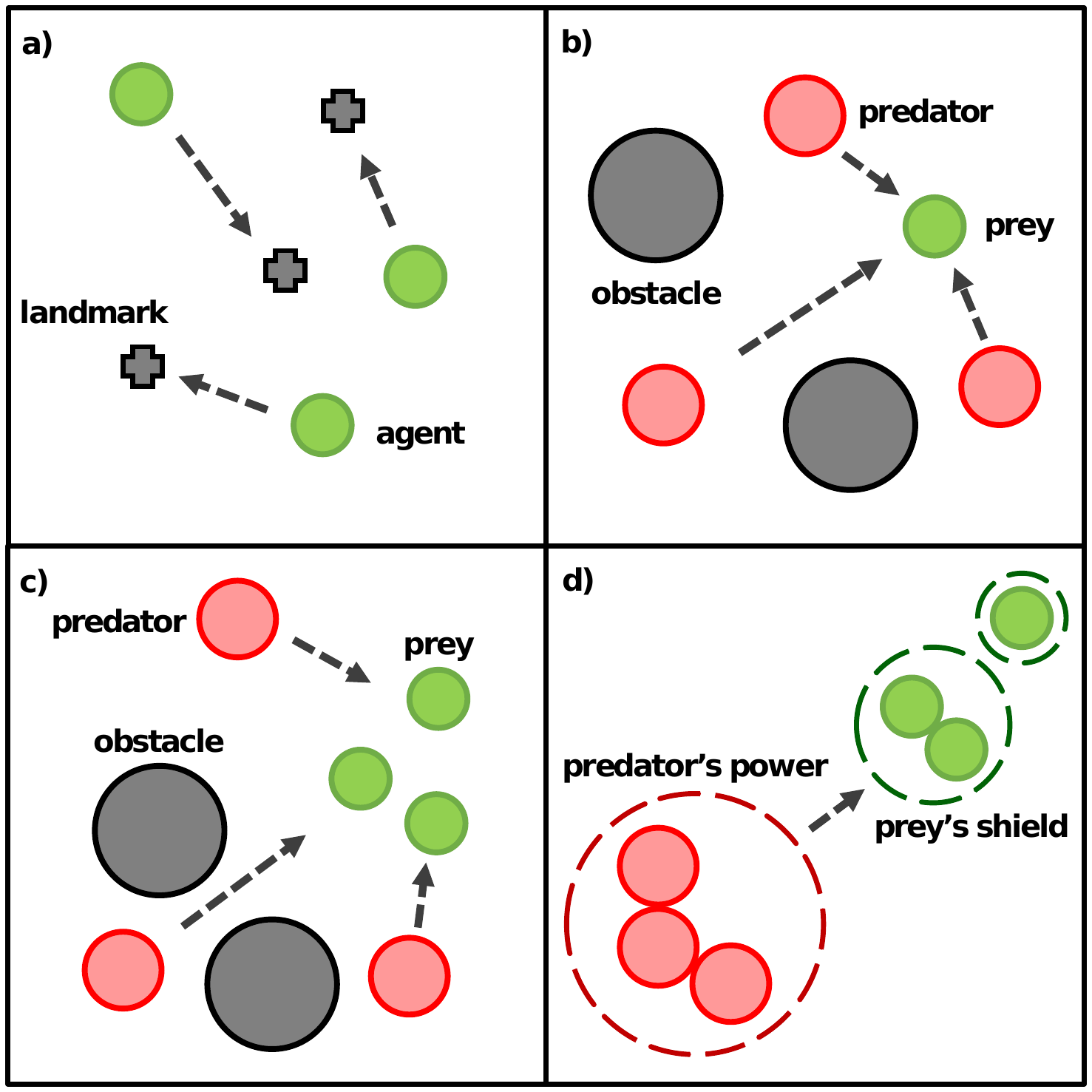}
  \caption{Illustrations of the experimental environments, including a) \emph{Cooperative Navigation}, b) \emph{3 vs. 1 Predator-Prey}, c) \emph{3 vs. 3 Predator-Prey}, and d) \emph{The More-The Stronger}.}
  \label{fig:fig2}
\end{figure}

\section{Experiments}
Figure \ref{fig:fig2} shows the environments we use to evaluate the performances of the proposed and baseline MARL algorithms. It includes cooperative environments that have been widely used in existing studies \cite{lowe2017multi,jiang2018learning} as well as mixed cooperative-competitive environments extended from well-known environments. As baseline algorithms for comparing the performances, we consider MADDPG and MAAC because they belong to learning-for-consensus approaches and are designed to process only local observation during the execution phase as HAMA does. These algorithms are more general for mixed cooperative-competitive games where communication is not always possible. %Note that the key difference between HAMA and the baselines is how to process local observations using relative inductive biases on the relationships among agents. 
For the cooperative navigation, we additionally consider ATOC, one of learning-to-communicate approaches, as a baseline because this game is fully cooperative; thus, the communication-based method can be naturally considered. All the performance measures are obtained by executing the trained policies with 3 different random seeds on 200 episodes.
% All the performance measures, the mean and standard deviation, are obtained by executing the trained policies on 3 sets of 200 episodes where each set starts with different random seeds.
Regarding the visual range of the agent, we assume that each agent observes up to three nearest neighboring agents per each group with relative positions and velocities in all the experiment settings.

\subsection{Cooperative Navigation}
First, the proposed model is validated in the cooperative navigation, where only cooperation among agents exists. 
In the game, all the agents, which are homogeneous, are required to reach one of the landmarks without colliding with each other. Each episode starts with $n$ randomly generated agents and landmarks and ends after 25 timesteps. During an episode, each agent receives $-d$, the distance to the nearest landmark, as a reward. In addition, each agent receives an additional reward, $-1$, whenever it collides with other agents during navigation. It is an optimal strategy that each agent occupies its distinct landmark.

\begin{figure}[t]
  \centering
  \includegraphics[width=0.38\textwidth]{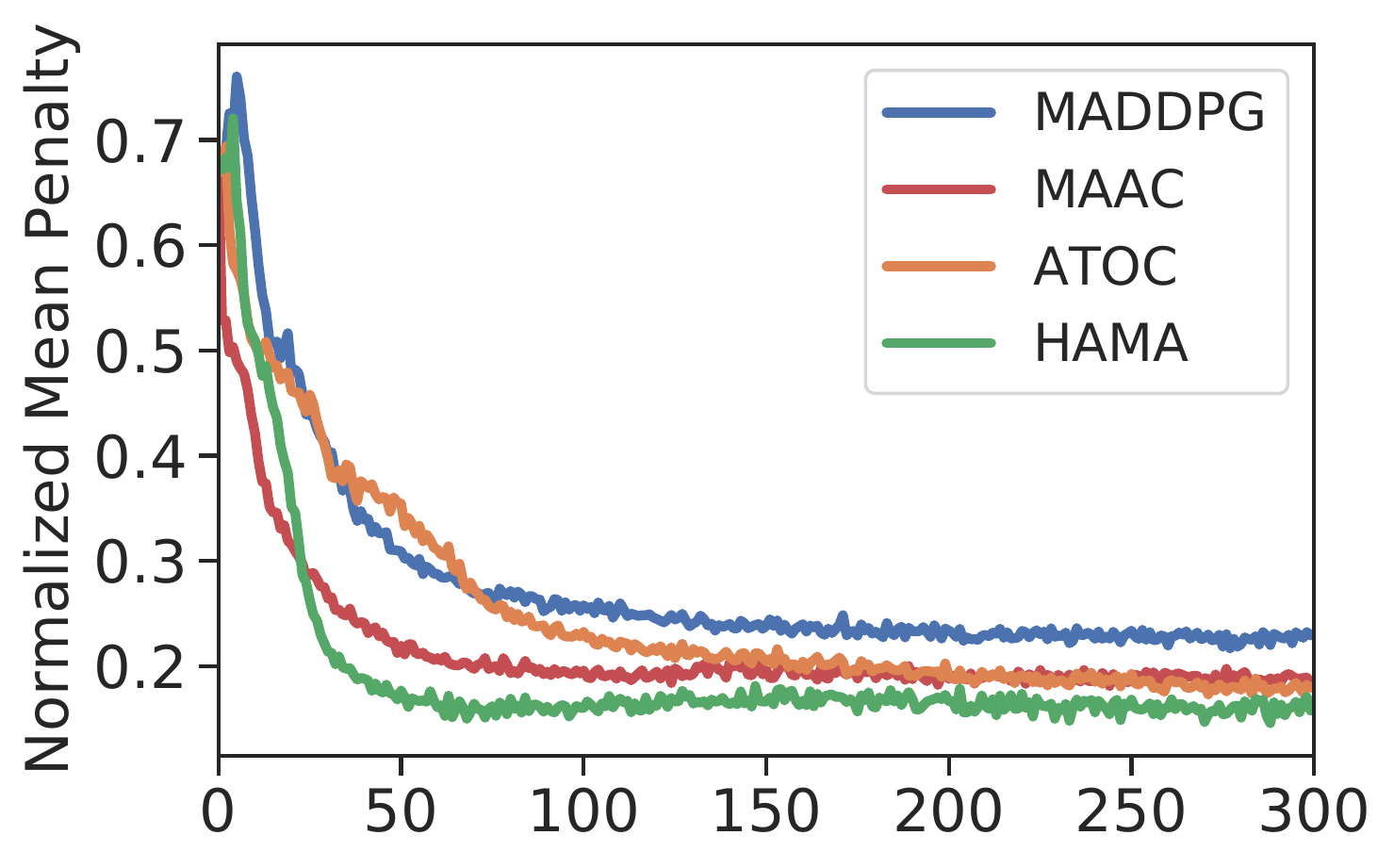}
  \caption{Penalties during training on cooperative navigation with 3 agents.}
  \label{fig:fig3}
\end{figure} 

\begin{table}[t]
  \caption{The mean and standard deviation of penalties in cooperative navigation.}
  \label{table:table1}
  \centering
  \begin{tabular}{ccccc}
    \toprule
    % \multicolumn{4}{c}{$n=3$}
    % \midrule
    % \multicolumn{1}{c}{\multirow{1}{2em}{n}}
     \multicolumn{1}{c}{$n$}&\multicolumn{1}{c}{MADDPG} & \multicolumn{1}{c}{MAAC}& \multicolumn{1}{c}{ATOC} & \multicolumn{1}{c}{HAMA} \\
    \midrule
    \multicolumn{1}{c}{3} & \multicolumn{1}{c}{0.22\scriptsize$\pm$0.010} & 0.19\scriptsize$\pm$0.012 &  0.17\scriptsize$\pm$0.007& \textbf{0.15}\scriptsize$\pm$0.008 \\    
    % \multicolumn{1}{c}{10} & \multicolumn{1}{c}{0.36\scriptsize$\pm$0.007} & 0.23\scriptsize$\pm$0.008 &  0.18\scriptsize$\pm$0.012& \textbf{0.10}\scriptsize$\pm$0.003 \\
    \midrule
    \multicolumn{5}{c}{Transfer learning using the policy trained with $n=3$}
    \\
    \midrule
    % \multicolumn{1}{c}{10} & \multicolumn{1}{c}{-} & - &  0.19\scriptsize$\pm$0.015& \textbf{0.13}\scriptsize$\pm$0.017 \\
    \multicolumn{1}{c}{50} & \multicolumn{1}{c}{-} & - &  0.09 (70)& \textbf{0.05} (\textbf{90}) \\
    \multicolumn{1}{c}{100} & \multicolumn{1}{c}{-} & - &  0.07 (81)& \textbf{0.04} (\textbf{96}) \\
    \bottomrule
  \end{tabular}
\end{table}

Figure \ref{fig:fig3} compares the normalized mean penalties of four different MARL algorithms (instead of representing results with rewards that are negative, we present the results in terms of penalty as a negative reward since it is more intuitive for this game). A smaller mean penalty implies closer to the nearest landmark and fewer collisions with other agents. The penalty is averaged over every 10,000 steps until reaching 3 million steps. 
% In the environment, we consider MADDPG, MAAC, and ATOC as the baselines.

% As a baseline, we consider MADDPG and MAAC because they are the most similar to our approach in that only local observation in the visual range is considered during execution. ATOC, one of learning-to-communicate approaches, is additionally considered in this experiment because all agents can be allowed to communicate with each other in the cooperative environment. 

In training, as shown in Figure \ref{fig:fig3}, HAMA converges fast to the lowest value. This is possible because HAMA effectively represents the state of each agent by considering their relative positions and velocities through HGAT. Table \ref{table:table1} compares the normalized mean penalties that the agents obtain during testing with 200 episodes with the trained models. ATOC achieves smaller mean penalty than MADDPG and MAAC. Because ATOC employs an active communication scheme based on attention network, it can effectively derive the cooperative behavior among agents. Our model has a lower mean penalty than ATOC. This indicates that the cooperative strategies trained by HAMA can effectively induce coordination among agents even without having active communication among agents.

%The transfer learning of Table \ref{table:table1} is the results of transferring the model trained in the environment of three agents directly to the environment of more agents. Since ATOC and HAMA have a transferable structure with shared actors between agents, their policies are transferable. HAMA outperforms ATOC in the penalty and the proportion of occupied landmarks in the environment with 50 and 100 smaller agents in a limited space.

Due to the use of a shared actor with efficient state representation, the trained policies by HAMA and ATOC can be applied to the cooperative game with any number of agents$/$landmarks, whereas the policies trained by MADDPG and MAAC cannot be transferred. The performance of transfer learning is also summarized in Table \ref{table:table1}. When the policies trained by 3 agents are used to play the game with 50 and 100 agents, HAMA has the lower average penalty and higher percentages of landmark occupation (provided in the parenthesis in the table) by the participating agents. Note that when we conduct the transfer learning experiments, we reduce the size of agents (25 times smaller) to have a large number of agents in the same environment, where each agent can observe three nearest agents and three landmarks.
% (visual range) with relative positions and velocities.

% The results of transferring in the environment of ten agents are almost the same as the result of direct learning. Furthermore, HAMA outperforms ATOC in the penalty and the proportion of occupied landmarks in the environment of 50 and 100 agents.

% In the environment of 50 and 100 agents, we use agents that are 25 times smaller than in the previous environment in order to scatter the agents in a given space.

% ATOC and HAMA achieve smaller penalties than MADDPG and MAAC. ATOC is a model designed for cooperative environments, and it derives cooperation through a communication channel with attention. Our model has a similar or smaller penalty than ATOC. This indicates that our model results in the cooperation among agents at a similar or better level than the model designed for cooperation. 

\begin{table}[t]
  \caption{The mean and standard deviation of scores for predators in 3 vs. 1 predator-prey game.} %for different counterpart models
  \label{table:table2}
  \centering
  \begin{tabular}{cccc}
    \toprule
    \multicolumn{1}{c}{predator} & \multicolumn{3}{c}{prey $(n=1)$}\\
    \cmidrule(r){2-4}
             \multicolumn{1}{c}{$(n=3)$}& \multicolumn{1}{c}{MADDPG} & \multicolumn{1}{c}{MAAC} & \multicolumn{1}{c}{HAMA} \\
    \midrule
    % \cmidrule(r){2-4}
    % \multirow{3}{3.5em}{predator \\$(n=3)$} &
    \multicolumn{1}{c}{MADDPG}  & 0.30\scriptsize$\pm$0.02 & 0.23\scriptsize$\pm$0.02&0.07\scriptsize$\pm$0.01 \\           
     \multicolumn{1}{c}{MAAC}       & 0.35\scriptsize$\pm$0.02 & \textbf{0.39}\scriptsize$\pm$0.03 &0.07\scriptsize$\pm$0.01     \\
     \multicolumn{1}{c}{HAMA}       & \textbf{0.45}\scriptsize$\pm$0.03 & \textbf{0.39}\scriptsize$\pm$0.03&\textbf{0.16}\scriptsize$\pm$0.01  \\
    \bottomrule
  \end{tabular}
\end{table}

\subsection{3 vs. 1 Predator-Prey}
A predator-prey game consists of two groups of agents competing with each other, along with obstacles that participate in the game but do not take an action. The goal of three homogeneous predators is to capture one prey, while the goal of the prey is to escape from the predators. For the predators to capture the prey, they need to cooperate with each other because of their slower speed and acceleration compared with those of the prey. Each predator gets a positive reward, $+10$, when it catches the prey, and the prey receives a negative reward, $-10$, when it is caught by a predator. When the prey leaves a certain zone, the prey receives a negative reward to prevent it from leaving farther. It is noteworthy that each agent seeks to maximize their accumulated rewards, which results in the competition between predators and prey.

We compare the performance of HAMA with two other models as baselines: MADDPG and MAAC.
% because they are designed for mixed cooperative-competitive environments.
Each model is trained while self-playing (i.e., predators and prey are trained with the same model), and the trained policies are validated while having the trained policies compete with other policies trained by different models. Table \ref{table:table2} summarizes the average scores that a predator can obtain per step in an episode. The results indicate that the predators trained by HAMA have higher or similar scores than MADDPG and MAAC when competing with the prey trained by other models. Similarly, the prey trained by HAMA results in the lowest score when competing with the predators trained by different models. Note that when HAMA plays the role of a single prey where no cooperation is required, it still performs best in defending itself from the predators because it effectively configures the relationships with the predators and obstacles by using HGAT.

\subsection{3 vs. 3 Predator-Prey}

The next game we consider is 3 vs. 3 predator-prey game, a variant of the original 3 vs. 1 predator-prey game. The game rules are identical to those of 3 vs. 1 predator-prey game. In this game, if a predator recaptures a prey that has already been captured, neither reward occurs. Instead, each predator receives an additional reward, $+10*t_r$, when the predators capture all preys, where $t_r$ is the number of remaining timesteps in the episode, and the game ends. Although the game is similar to that of the original predator-prey game, the optimal strategy of the agents is no longer clear because more diverse and complex strategic interactions occur among the two groups of agents. For example, a predator can choose to either cooperate with other predators to chase a prey or to capture a prey individually if the prey is nearby. In addition to MADDPG and MAAC, we consider two heuristic strategies for the predators. In Heuristic 1, all the predators chase the same prey that has not been captured yet. In Heuristic 2, each predator chases the prey closest to the predator.

\begin{table}[t]

    \caption{The mean and standard deviation of scores for predators in 3 vs. 3 predator-prey game.} 
    \label{table:table3}
    \centering
  \begin{tabular}{cccc}
    \toprule

    \multicolumn{1}{c}{predator} & \multicolumn{3}{c}{prey $(n=3)$}\\
    \cmidrule(r){2-4}
    
    \multicolumn{1}{c}{$(n=3)$}& \multicolumn{1}{c}{MADDPG} & \multicolumn{1}{c}{MAAC} & \multicolumn{1}{c}{HAMA}\\
    \midrule
    \multicolumn{1}{c}{Heuristic1} & 0.35\scriptsize$\pm$0.07 & 0.15\scriptsize$\pm$0.10 & 0.005\scriptsize$\pm$0.001\\
     \multicolumn{1}{c}{Heuristic2}  & 0.72\scriptsize$\pm$0.10 & 0.30\scriptsize$\pm$0.14&0.01\scriptsize$\pm$0.001\\
      \multicolumn{1}{c}{MADDPG}  & 1.18\scriptsize$\pm$0.13 & 1.05 \scriptsize$\pm$0.22 &0.02\scriptsize$\pm$0.01\\           
      \multicolumn{1}{c}{MAAC}   & 0.65\scriptsize$\pm$0.20 & 0.33\scriptsize$\pm$0.13 & 0.07\scriptsize$\pm$0.04\\
      \multicolumn{1}{c}{HAMA}    & \textbf{6.33}\scriptsize$\pm$0.10 & \textbf{3.36}\scriptsize$\pm$0.34 &\textbf{1.19}\scriptsize$\pm$0.09\\

    \bottomrule
  \end{tabular}

\end{table}

\begin{table}[t]

  \caption{The mean and standard deviation of scores for different architectures in 3 vs. 3 predator-prey game.}
  \label{table:table4}
  \centering
  \begin{tabular}{cc}
    \toprule
    \multicolumn{1}{c}{predator} & \multicolumn{1}{c}{prey $(n=3)$}\\
    \cmidrule(r){2-2}
    \multicolumn{1}{c}{$(n=3)$}& \multicolumn{1}{c}{HG-IAGA (HAMA)} \\
    \midrule
    % \multirow{5}{4em}{predator \\$(n=3)$} &
    \multicolumn{1}{l}{SG-IAA (MAAC)}  & 0.07\scriptsize$\pm$0.04  \\           
   \multicolumn{1}{l}{HG-NA}       & 0.57\scriptsize$\pm$0.07  \\
   \multicolumn{1}{l}{HG-IAA}       &  1.03\scriptsize$\pm$0.08 \\
   \multicolumn{1}{l}{HG-IGA}       & 0.37\scriptsize$\pm$0.06  \\
   \multicolumn{1}{l}{HG-IAGA (HAMA)}       & \textbf{1.19}\scriptsize$\pm$0.09  \\
    \bottomrule
  \end{tabular}
 \end{table}

Table \ref{table:table3} compares the results of the game when the predators and preys, each of which is trained by self-playing, compete against each other. As shown in the table, the predators trained by HAMA achieve the highest scores against the preys trained by all other algorithms. Similarly, the preys trained by HAMA defend the best against the predators trained by all different algorithms, including the two heuristic strategies. The performance of HAMA is incomparably superior to those of other methods in both roles of predator and prey. This is remarkable in that HAMA and MAAC achieve a similar performance in the 3 vs. 1 predator-prey game where the only strategy of the predators is to cooperate to capture a unique prey. Meanwhile, in the 3 vs. 3 predator-prey game, each predator can choose from various strategies, such as cooperating with other predators or chasing prey individually.
% For example, one predator can choose between cooperating with other predators or chasing prey individually.
When chasing prey, a predator can also choose which prey to chase. The superior performance of HAMA is possible because it learns to represent better the hierarchical relationships among agents in the dynamic game owing to the relative inductive biases imposed by HGAT.

We validate our hypothesis on the success of HAMA's strategy by conducting an ablation study. Table \ref{table:table4} summarizes the performances of the following variant models:
\begin{itemize}%[noitemsep,topsep=0pt]
  \item SG-IAA:  Single-Graph \& Inter-Agent Attention
  \item HG-NA:  Hierarchical-Graph \& No Attention
  \item HG-IAA:  Hierarchical-Graph \& Inter-Agent Attention
  \item HG-IGA:  Hierarchical-Graph \& Inter-Group Attention
  \item HG-IAGA (HAMA):  Hierarchical-Graph \& Inter-Agent \& Inter-Group Attention
\end{itemize}

Note that the SG-IAA has a similar architecture with MAAC.
% except the difference that MAAC uses this architecture for only representing the critic.
Compared to the SG-IAA, a hierarchical graph attention architecture always scores higher regardless of whether attention is used. We assume that this effectiveness is due to the use of enhanced relative inductive biases
% , represented by HGAT in HAMA,
regarding both the agent-level interactions and the group-level interactions. In comparing the role of attention, when both attentions are considered, the HG-IAGA outperforms the others. The combination of hierarchical graph structure and specially designed attentions is a key factor that induces the superior performance of HAMA.

\subsubsection{Transfer Learning.}
\begin{figure}[t]
  \centering
  \includegraphics[width=0.35\textwidth]{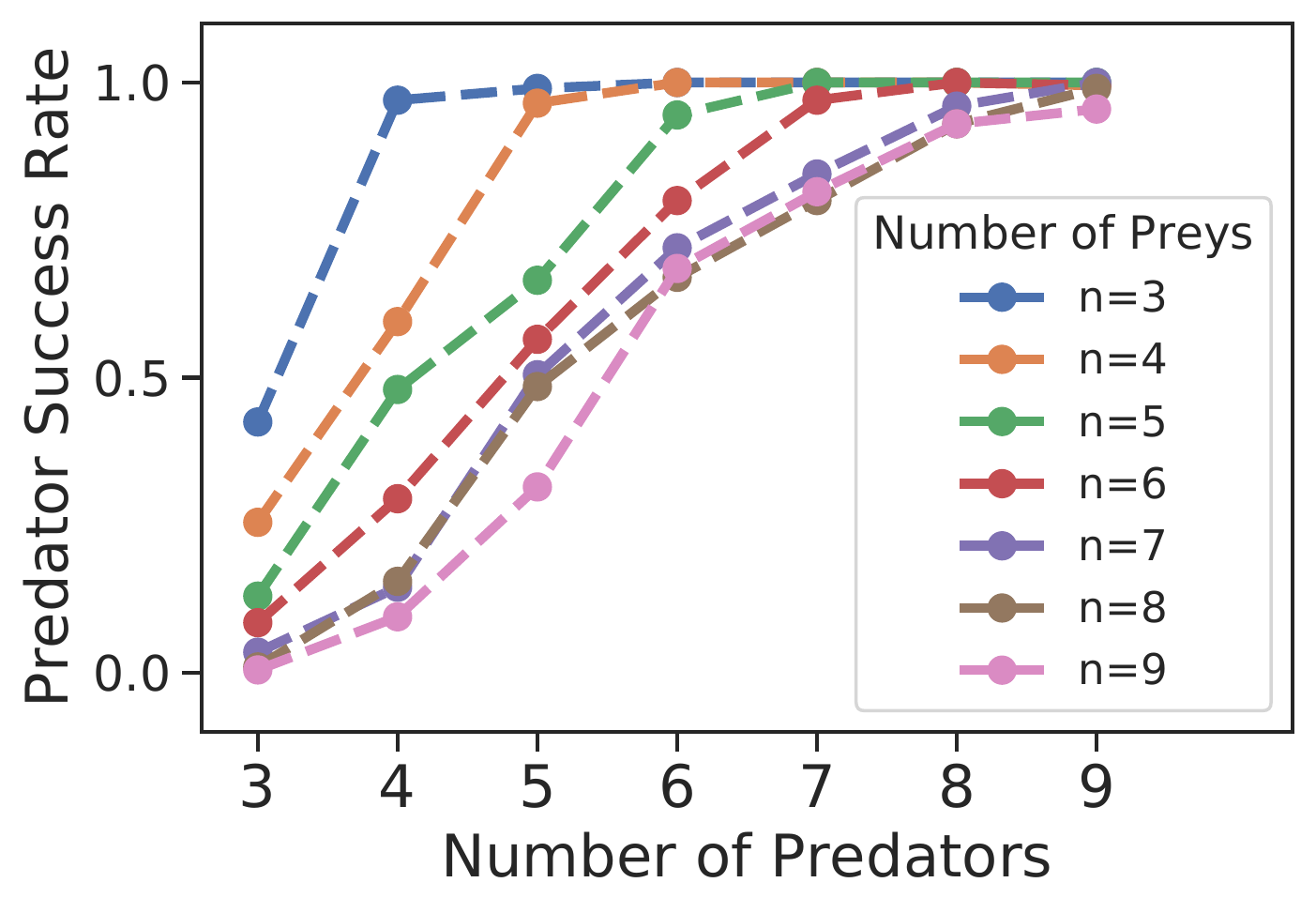}
  \caption{Results of transfer learning with the trained policies by HAMA in 3 vs. 3 predator-prey game. The success rate of predators is the rate of episodes in which predators capture all preys.}
  \label{fig:fig4}
\end{figure} 

In general, when the number of predators is large and the number of preys is small, the predators
% as a team
have a higher chance to win the game (i.e., capture all the preys within a single episode). As shown in Figure \ref{fig:fig4}, this general trend is well realized when the predator and prey policies trained by HAMA in the 3 vs. 3 predator-prey game are transferred to play an $m$ vs. $n$ predator-prey game.
% (zero-shot learning).
It shows that the success rate of the predators is close to 1 when $m$ (number of predators) $> n$ (number of preys). %The capability of such transfer learning is owing to the use of the graph neural network combined with attentions. This combination allows the policies to learn complex interactions among agents from pairwise-agent and pairwise-group interactions.

\subsubsection{Interpreting Strategies.}
\begin{figure}[t]
  \centering
  \includegraphics[width=0.4\textwidth]{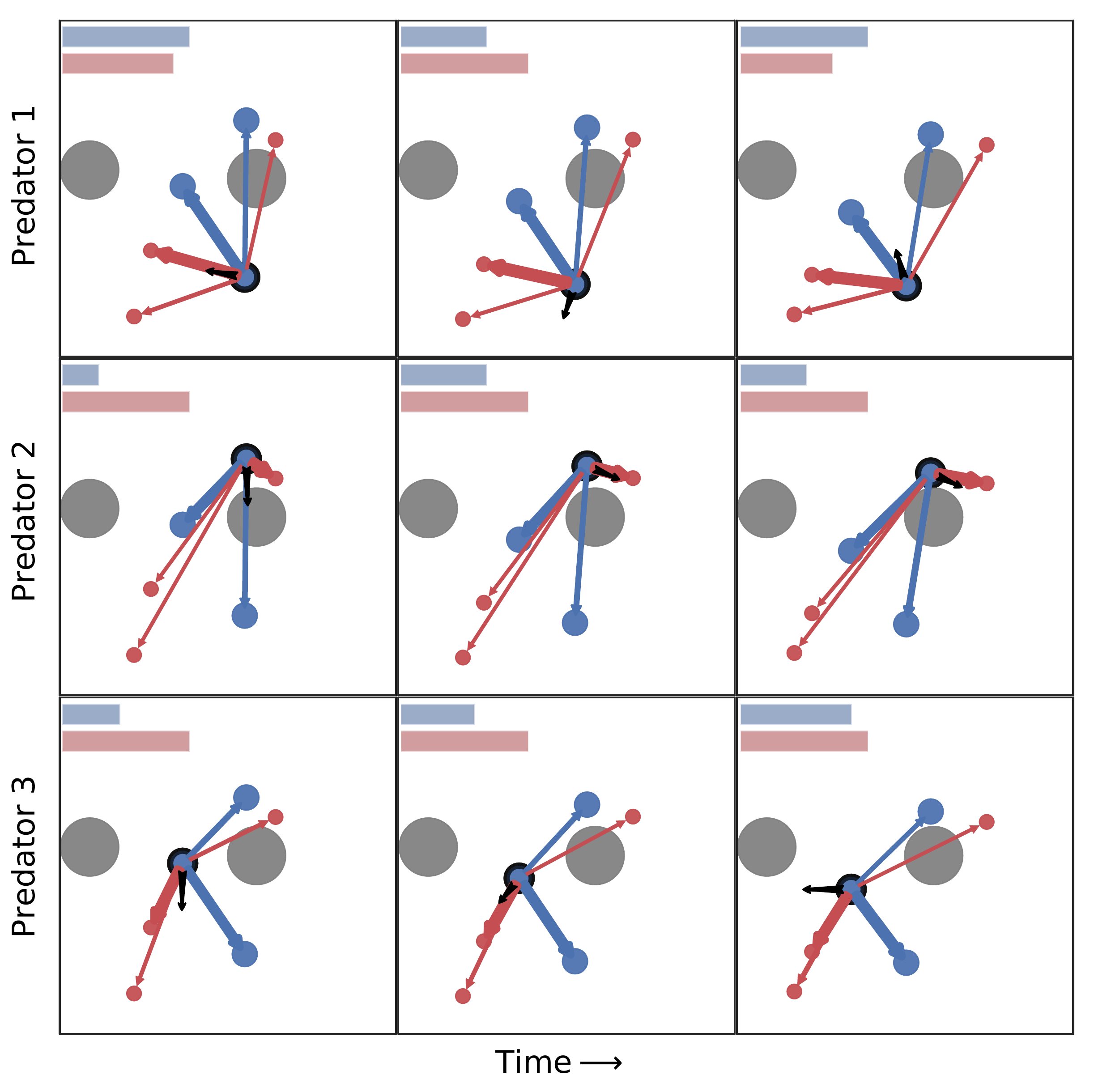}
  \caption{Reasoning on the strategy of HAMA.}
  \label{fig:fig5}
\end{figure} 

We explain why a certain action of the agent is induced at a certain state by analyzing and interpreting the inter-agent and inter-group attention weights in HAMA.
In Figure \ref{fig:fig5}, the blue, red, and gray circles represent the predators, preys, and obstacles, respectively. The plots in each row show how each predator agent, which is represented by the blue circle with a black outline, attends other agents in the same and different groups over time. The width of the arrow indicates the magnitude of the attention weight $\alpha_{ij}^k$ on the agent the arrow is pointing out in each group. The blue and red bars at the top of each figure indicate the magnitudes of inter-group attention weights $\beta_i^k$ to the predator ($k=1$) and the prey groups ($k=2$), respectively. The black arrow indicates the agent's action (i.e., direction and speed). From the predator's perspective, the attention to predators and preys can be interpreted as the attention to cooperation and competition. Figure \ref{fig:fig5} depicts the situation where predator $1$ and $3$ increase the cooperative attention (i.e., attention to the same group) over time to jointly chase the prey into a corner of the box. Meanwhile, predator $2$ attempts to catch one prey, whose strategy is indicated by the attention to the competition (i.e., attention to the different group).
% More various figures on strategy interpretation are included in the supplemental material.

\subsection{The More-The Stronger}
% \begin{table}[t]
%   \caption{The mean and standard deviation of scores of predators on the more-the stronger game.}
%   \label{table:table5}
%   \centering
%     \begin{tabular}{cccc}
%         \toprule
    
%         \multicolumn{1}{c}{predator} & \multicolumn{3}{c}{prey $(n=3)$}\\
%         \cmidrule(r){2-4}
        
%         \multicolumn{1}{c}{$(n=3)$} & \multicolumn{1}{c}{MADDPG} & \multicolumn{1}{c}{MAAC} & \multicolumn{1}{c}{HAMA}\\
%         \midrule
%         % \multirow{5}{4em}{predator \\$(n=3)$} &
%         \multicolumn{1}{c}{Heuristic1} & 0.19\scriptsize$\pm$0.05 & 0.17\scriptsize$\pm$0.09 & 0.003\scriptsize$\pm$0.001\\
%         \multicolumn{1}{c}{Heuristic2}  & 0.73\scriptsize$\pm$0.10 & 0.49\scriptsize$\pm$0.18&0.005\scriptsize$\pm$0.001\\
%         \multicolumn{1}{c}{MADDPG}  & 1.64\scriptsize$\pm$0.13 & 1.81\scriptsize$\pm$0.29 &0.02\scriptsize$\pm$0.001\\      \multicolumn{1}{c}{MAAC}   & 0.77\scriptsize$\pm$0.22 & 0.71\scriptsize$\pm$0.23 & 0.64\scriptsize$\pm$0.20\\
%         \multicolumn{1}{c}{HAMA}    & \textbf{5.45}\scriptsize$\pm$0.12 & \textbf{3.45}\scriptsize$\pm$0.30 &\textbf{2.08}\scriptsize$\pm$0.13\\
    
%         \bottomrule
%     \end{tabular}
% \end{table}

The more-the stronger game keeps the framework of the 3 vs. 3 predator-prey game. The additional game rule in this game is that when the preys are clustered together, only a group of predators whose size is equal to or larger than that of the clustered preys can capture the preys. For example, one predator can capture one prey by itself, but three-gathered predators are required to capture three-gathered preys. HAMA outperforms other models in this game.% The results are summarized in supplemental material. %HAMA outperforms other models in this game as well.  

\section{Conclusions}
We herein proposed a multi-agent actor-critic model based on state representation by a hierarchical graph attention network. Empirically, we demonstrated that the learned model outperformed other MARL models on a variety of cooperative and competitive multi-agent environments. In addition, the proposed model has been proven to facilitate the transfer of learned policies to new tasks with different agent compositions and allow one to interpret the learned strategies.

\section{Acknowledgments}
This work was supported by the National Research Foundation of Korea (NRF) grant funded by the Korea government (MSIT) (No. 2017R1A2B4006290) and by the Technology Innovation Program (or Industrial Strategic Technology Development Program, 10067705, Development of AI-based Smart Construction System for 20\% Cost Saving) funded by the Ministry of Trade, Industry \& Energy (MI, Korea).

% \bibliography{AAAI-RyuH.3962}

\bibliographystyle{aaai}

\begin{figure*}[h!]
    \label{fig:fig7}
    \caption{Reasoning on the strategy of HAMA in another episode.}
    \subfigure[]{\label{fig:a}\includegraphics[width=46mm]{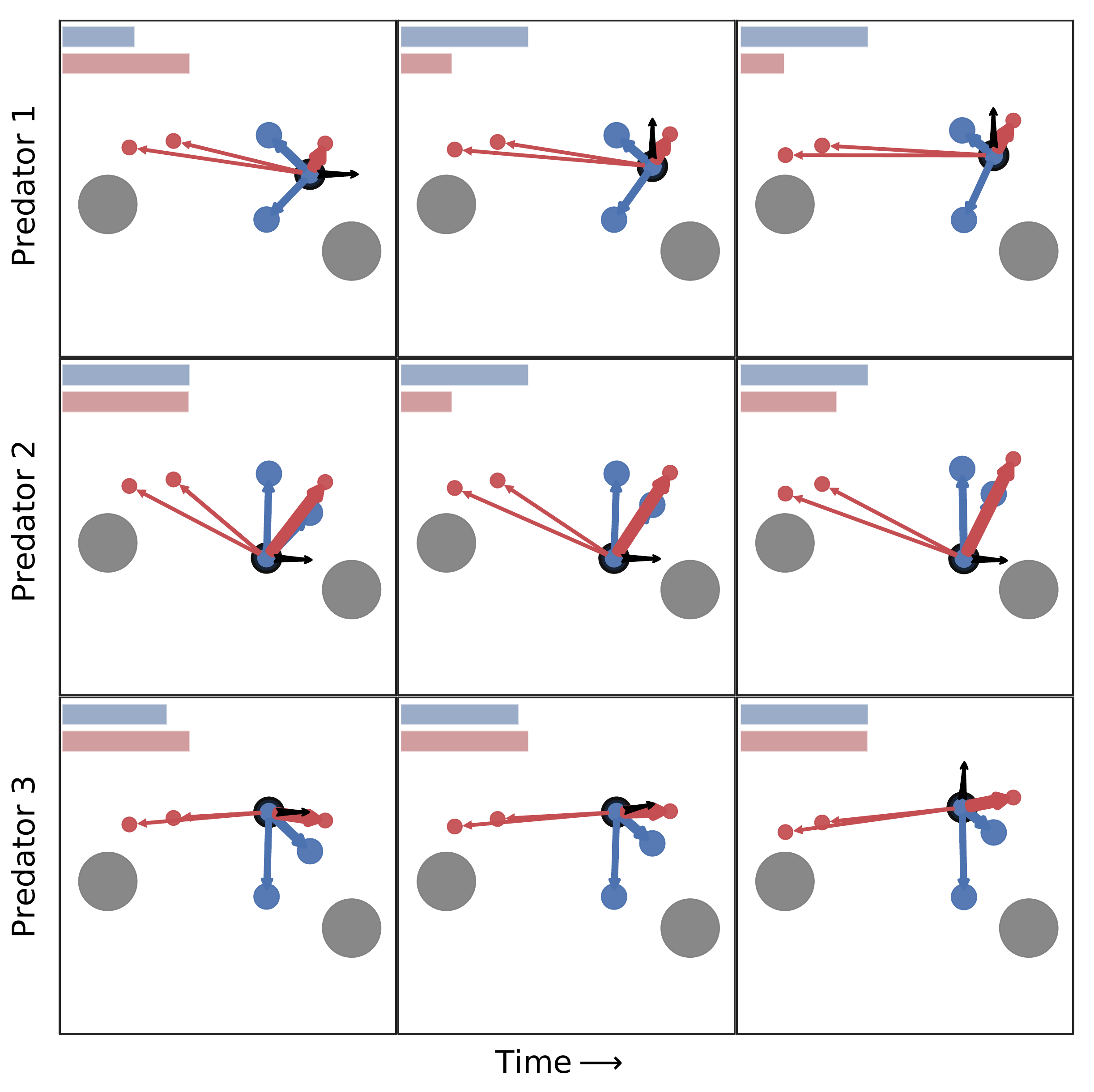}}
    \hspace{.11\textwidth}
    \subfigure[]{\label{fig:b}\includegraphics[width=46mm]{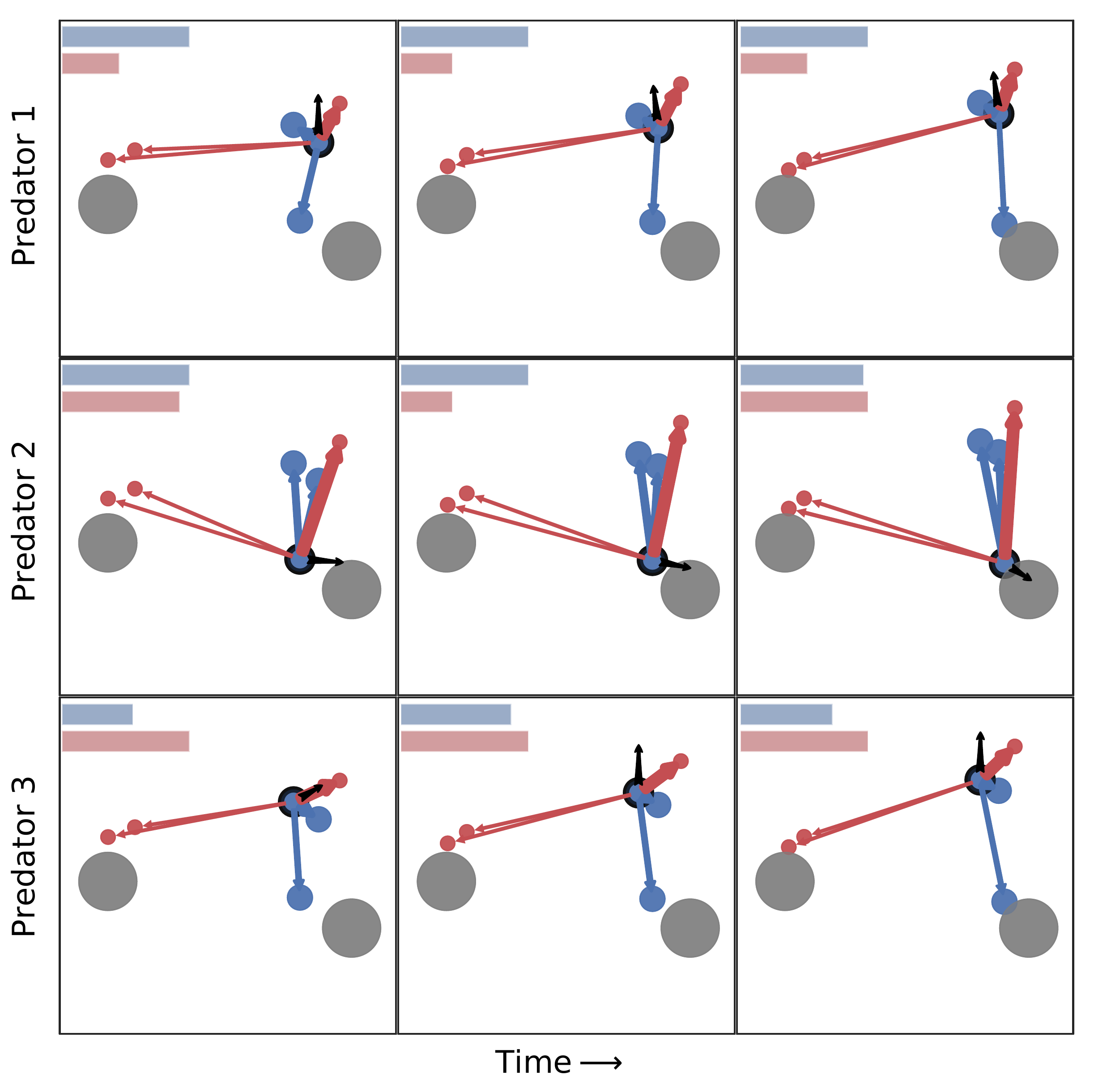}}
    \hspace{.11\textwidth}
    \subfigure[]{\label{fig:c}\includegraphics[width=46mm]{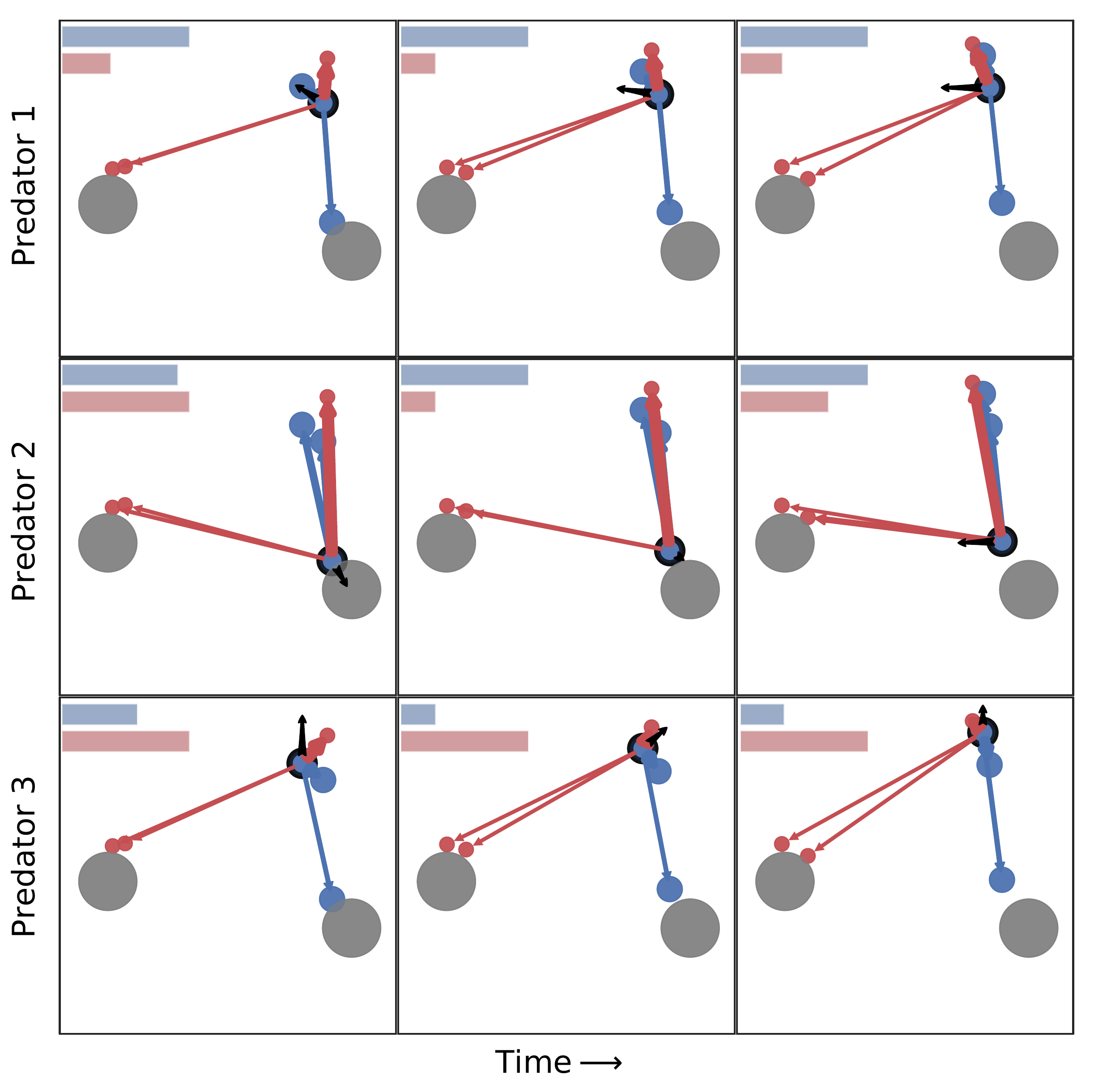}}
    \hspace{.11\textwidth}
    \subfigure[]{\label{fig:d}\includegraphics[width=46mm]{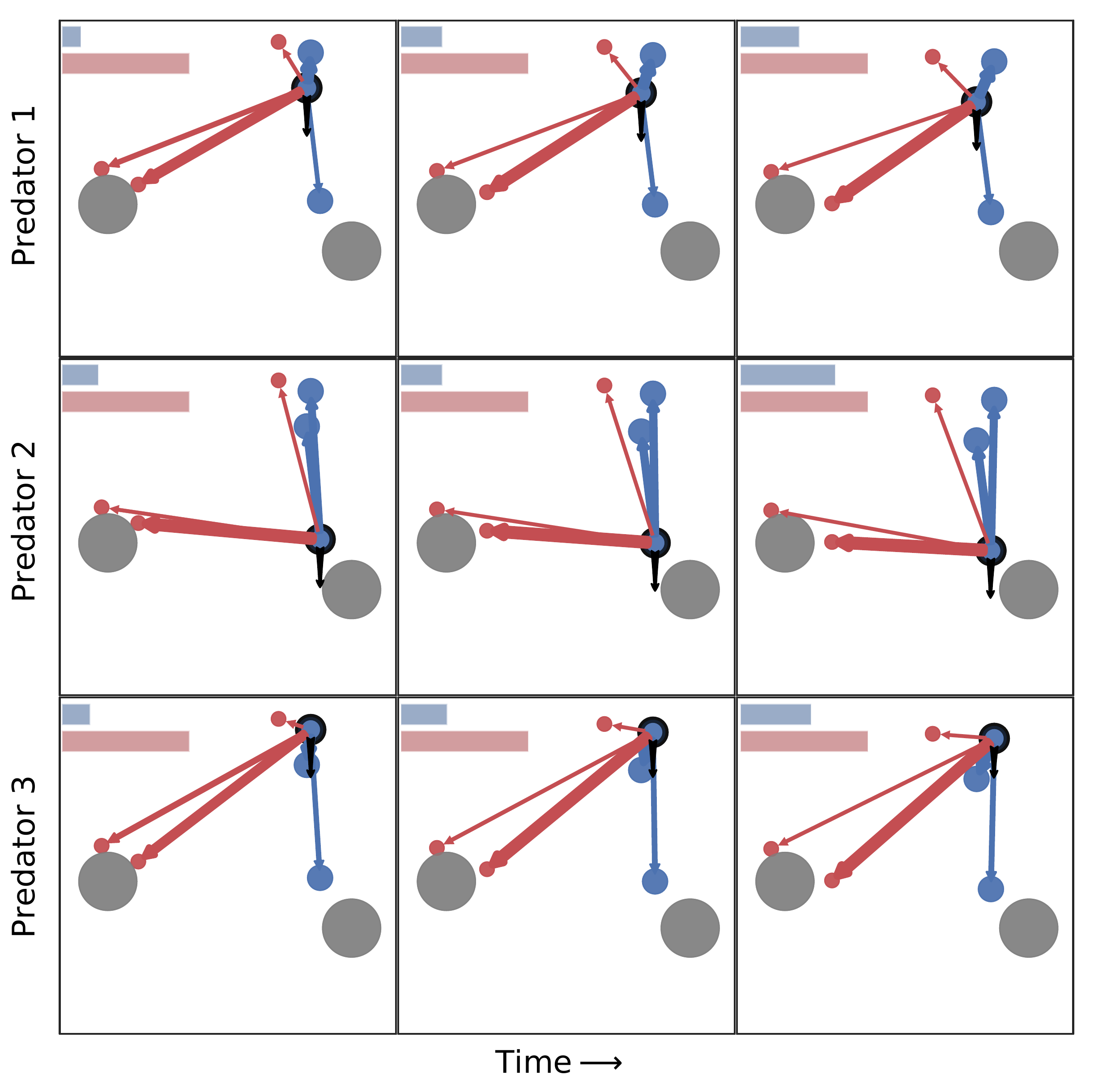}}
    \hspace{.11\textwidth}
    \subfigure[]{\label{fig:e}\includegraphics[width=46mm]{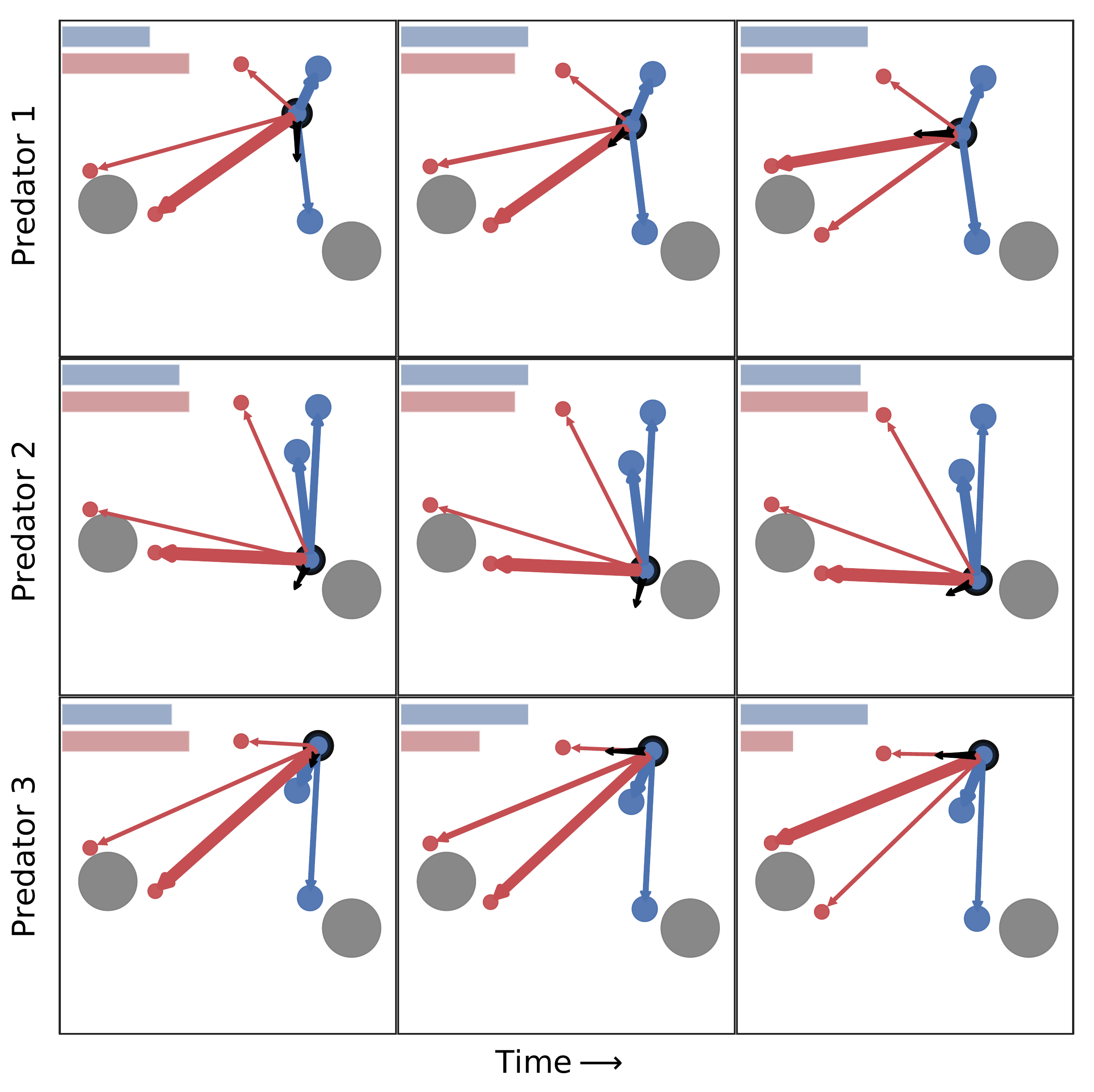}}
    \hspace{.11\textwidth}
    \subfigure[]{\label{fig:f}\includegraphics[width=46mm]{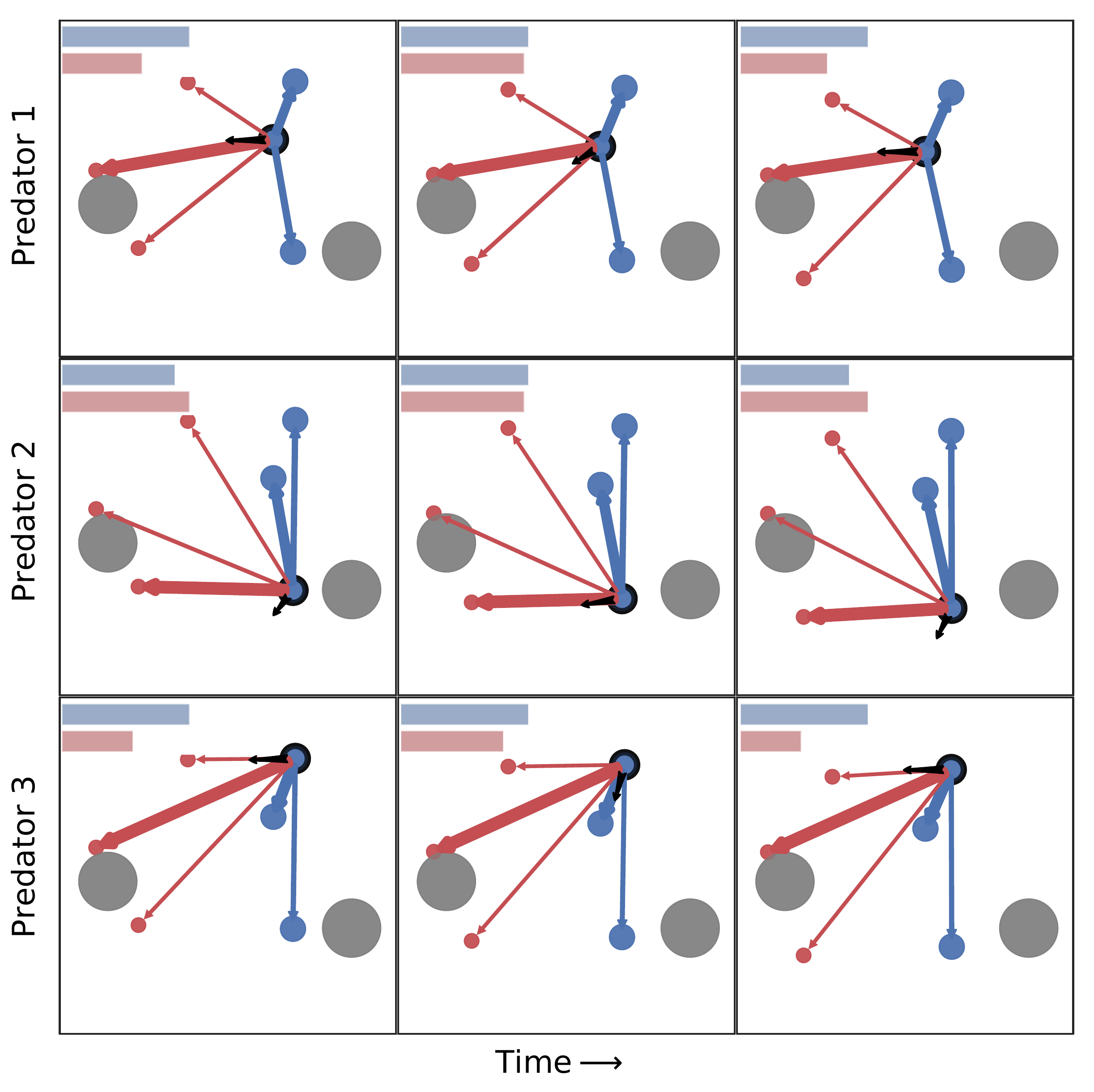}}
    \hspace{.11\textwidth}
    \subfigure[]{\label{fig:g}\includegraphics[width=46mm]{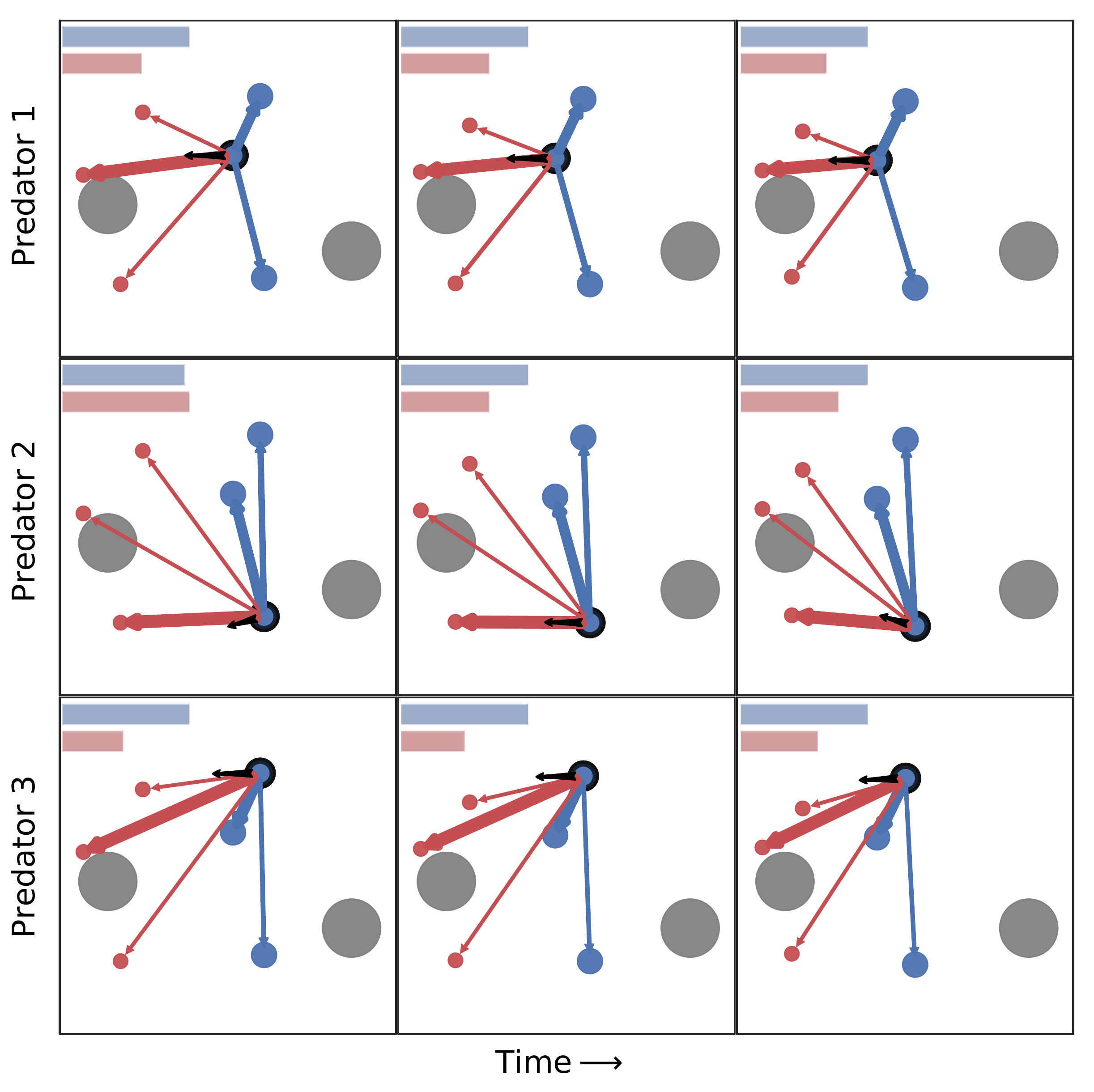}}
    \hspace{.11\textwidth}
    \subfigure[]{\label{fig:h}\includegraphics[width=46mm]{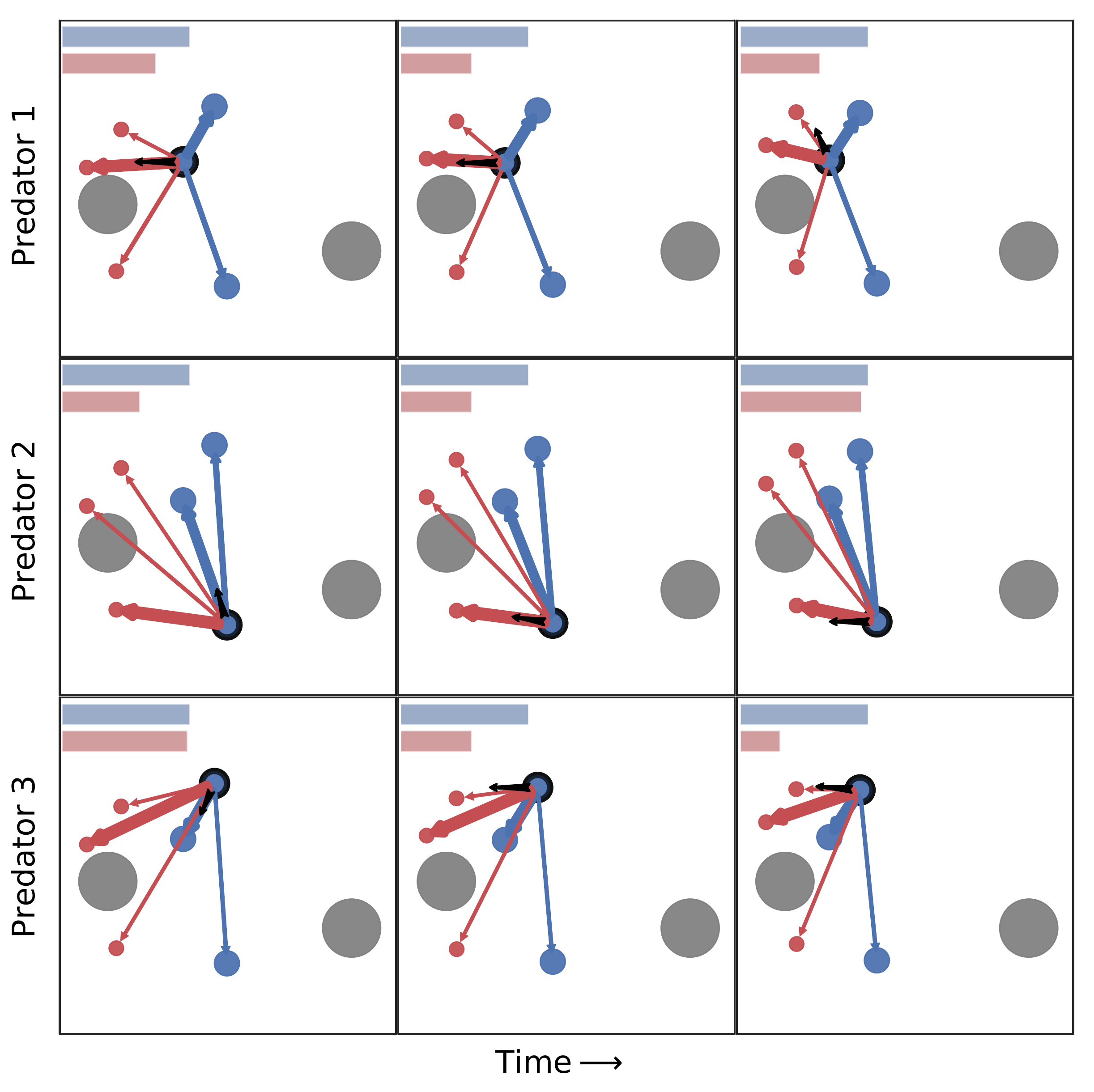}}
    \hspace{.11\textwidth}
    \subfigure[]{\label{fig:i}\includegraphics[width=46mm]{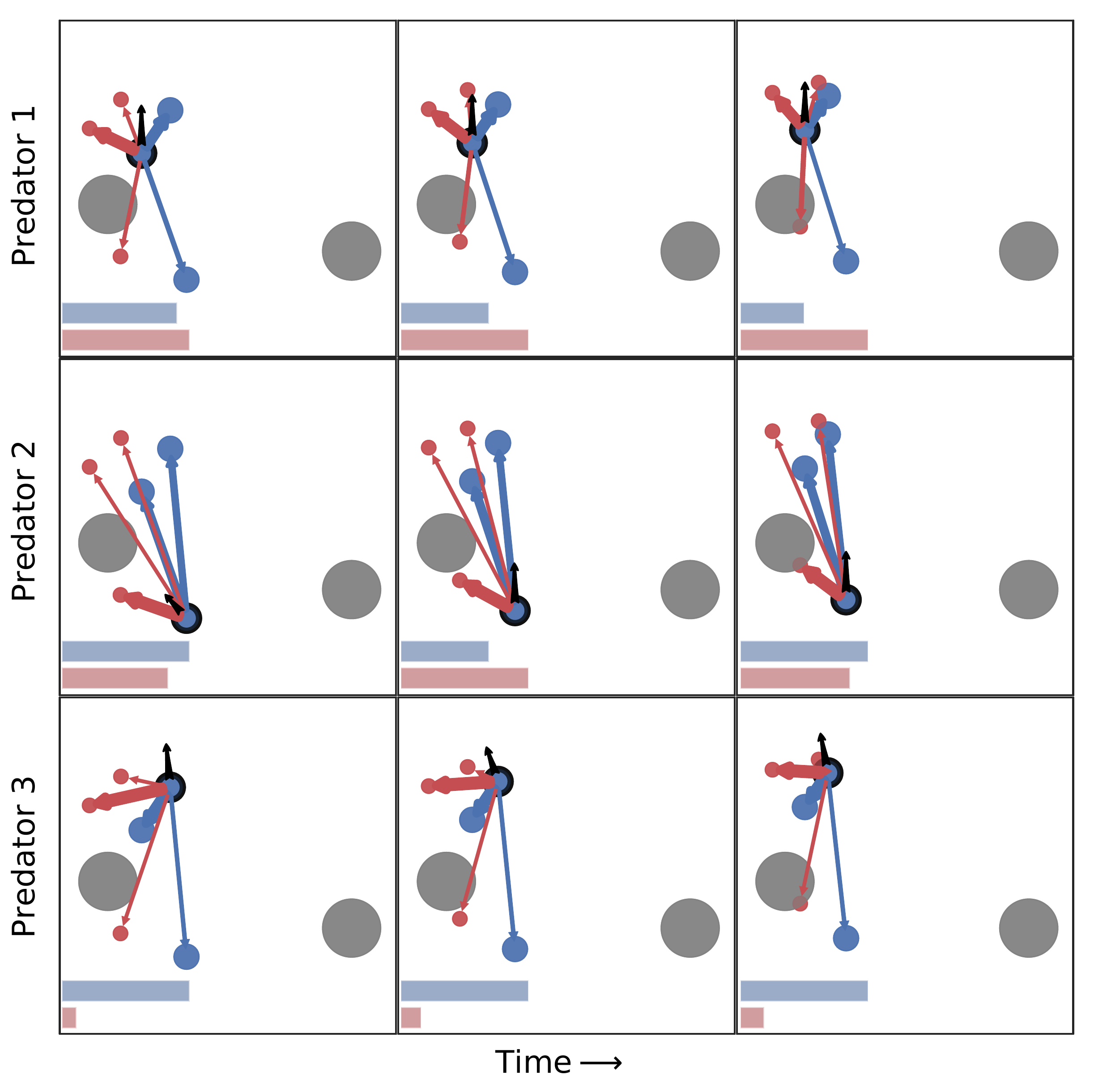}}
    \hspace{.11\textwidth}
    \subfigure[]{\label{fig:j}\includegraphics[width=46mm]{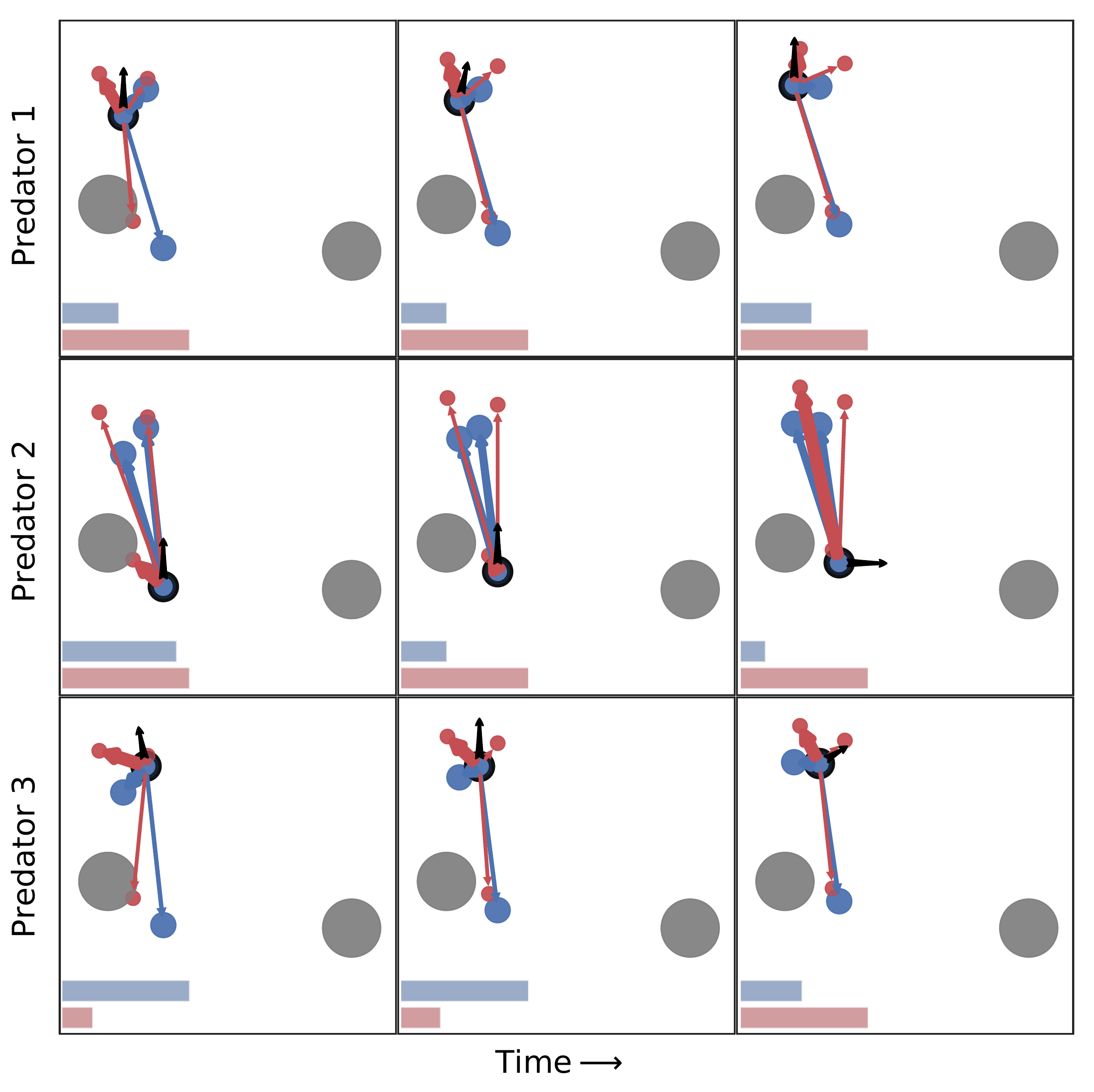}}
    \hspace{.11\textwidth}
    \subfigure[]{\label{fig:k}\includegraphics[width=46mm]{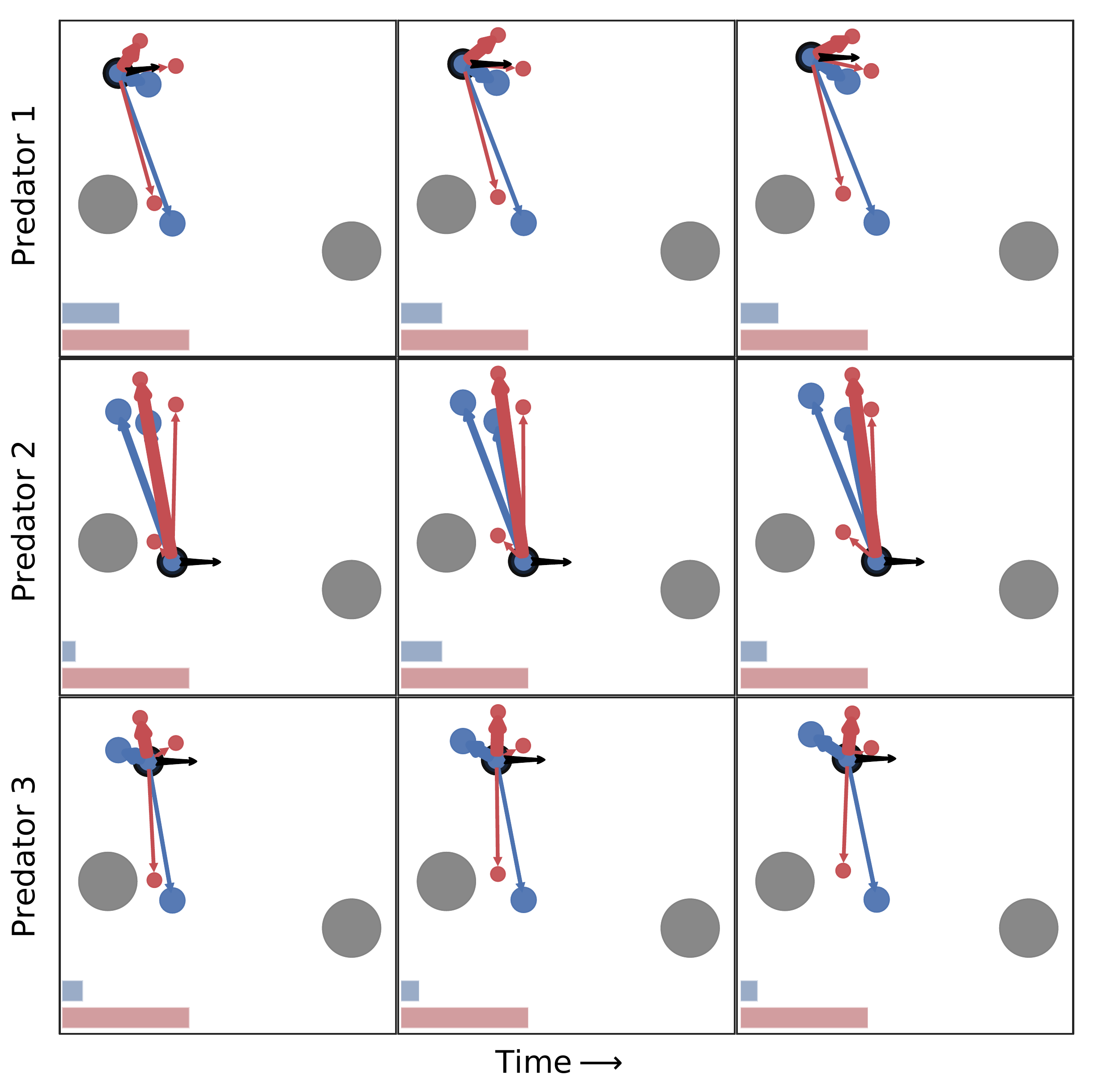}}
    \hspace{.11\textwidth}
    \subfigure[]{\label{fig:l}\includegraphics[width=46mm]{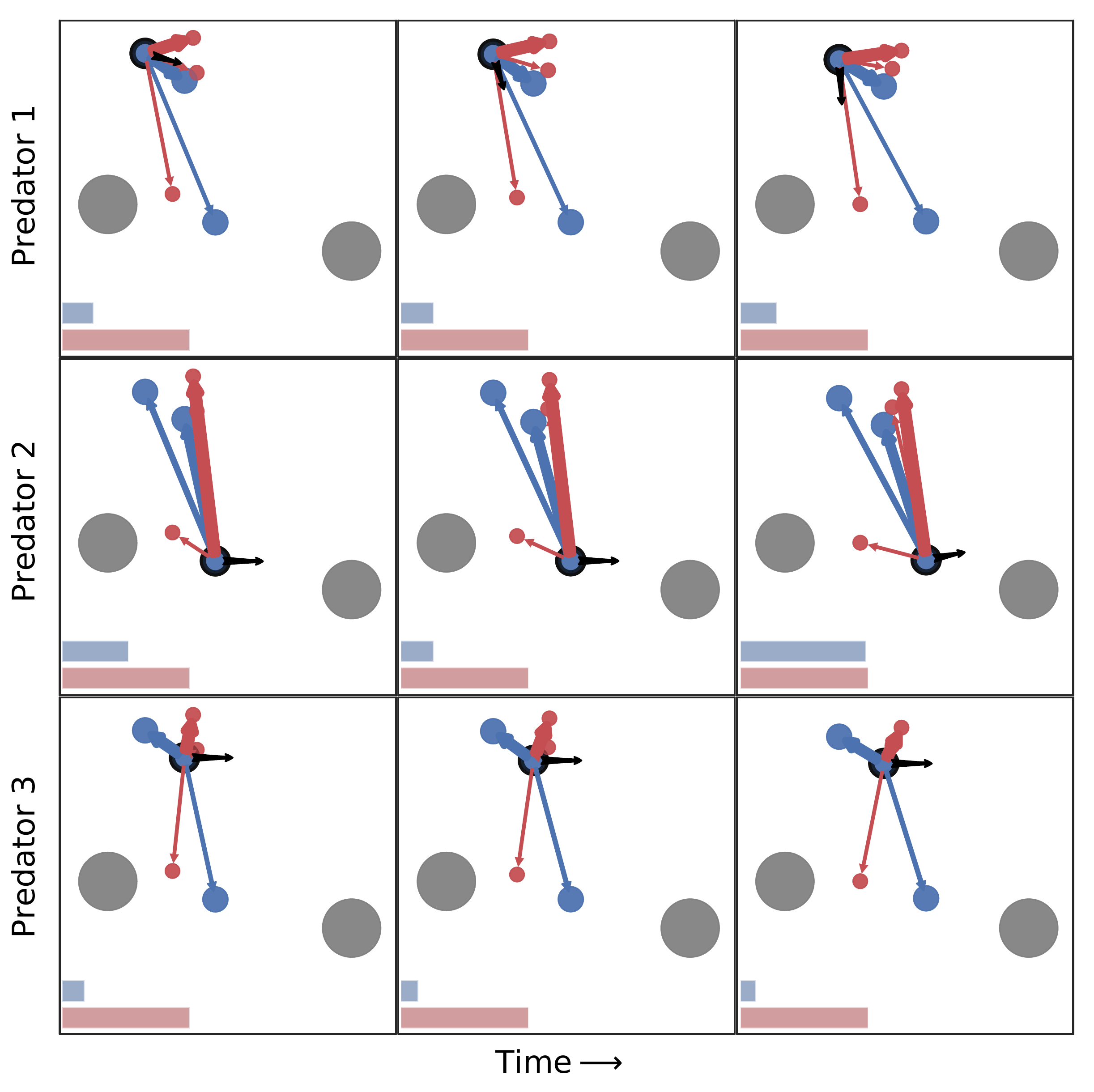}}
\end{figure*}
% \newpage
\begin{figure*}[h]
    % \subfigure[]{\label{fig:g}\includegraphics[width=46mm]{attention_anl_re_seed1_7.pdf}}
    % \subfigure[]{\label{fig:h}\includegraphics[width=46mm]{attention_anl_re_seed1_8.pdf}}
    % \subfigure[]{\label{fig:i}\includegraphics[width=46mm]{attention_anl_re_seed1_9.pdf}}
    % \subfigure[]{\label{fig:j}\includegraphics[width=46mm]{attention_anl_re_seed1_10.pdf}}
    % \hspace{.125\textwidth}
    % \subfigure[]{\label{fig:k}\includegraphics[width=46mm]{attention_anl_re_seed1_11.pdf}}
    % \hspace{.125\textwidth}
    % \subfigure[]{\label{fig:l}\includegraphics[width=46mm]{attention_anl_re_seed1_12.pdf}}
    % \hspace{.125\textwidth}
    \subfigure[]{\label{fig:m}\includegraphics[width=46mm]{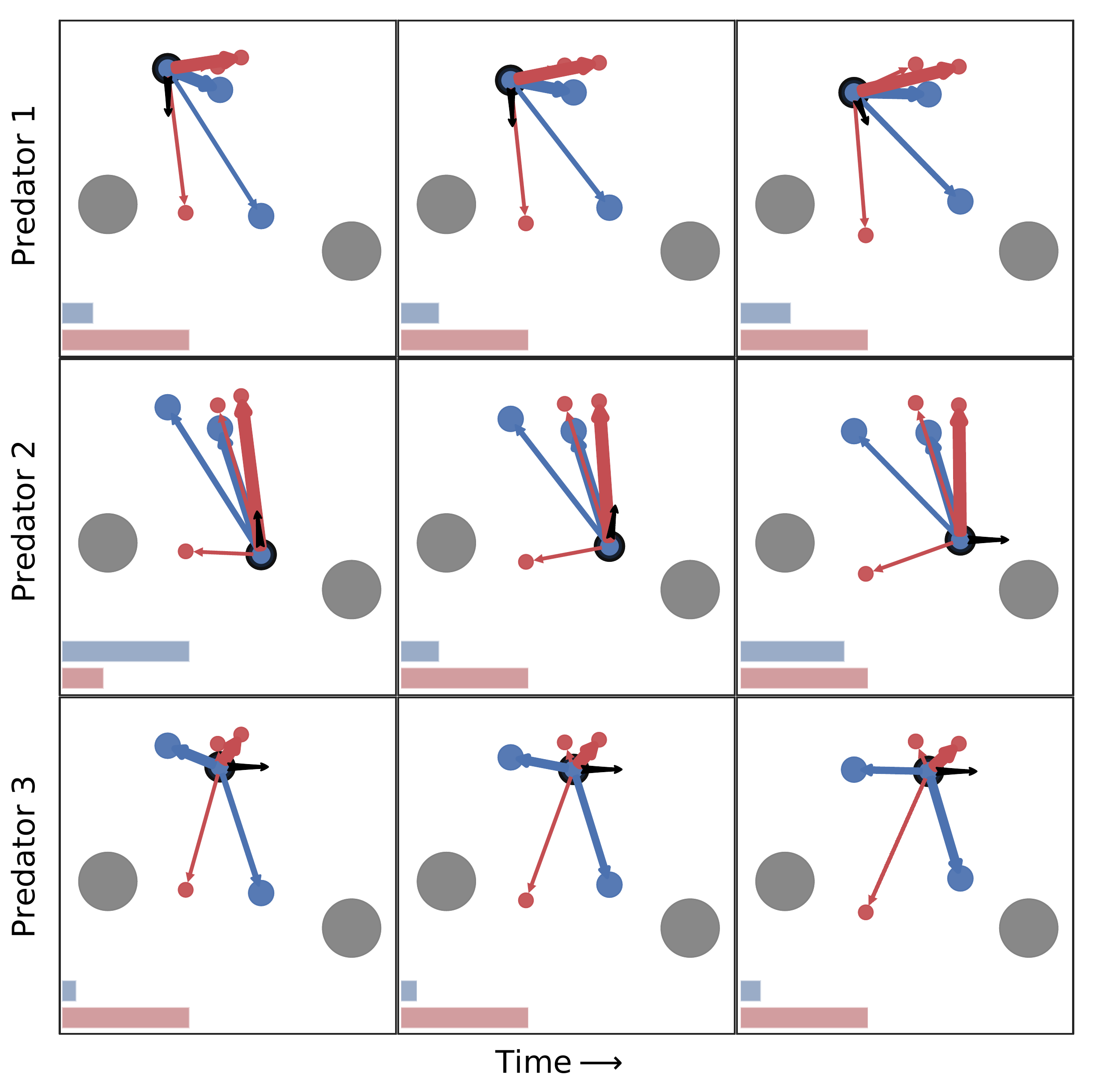}}
    \hspace{.11\textwidth}
    \subfigure[]{\label{fig:n}\includegraphics[width=46mm]{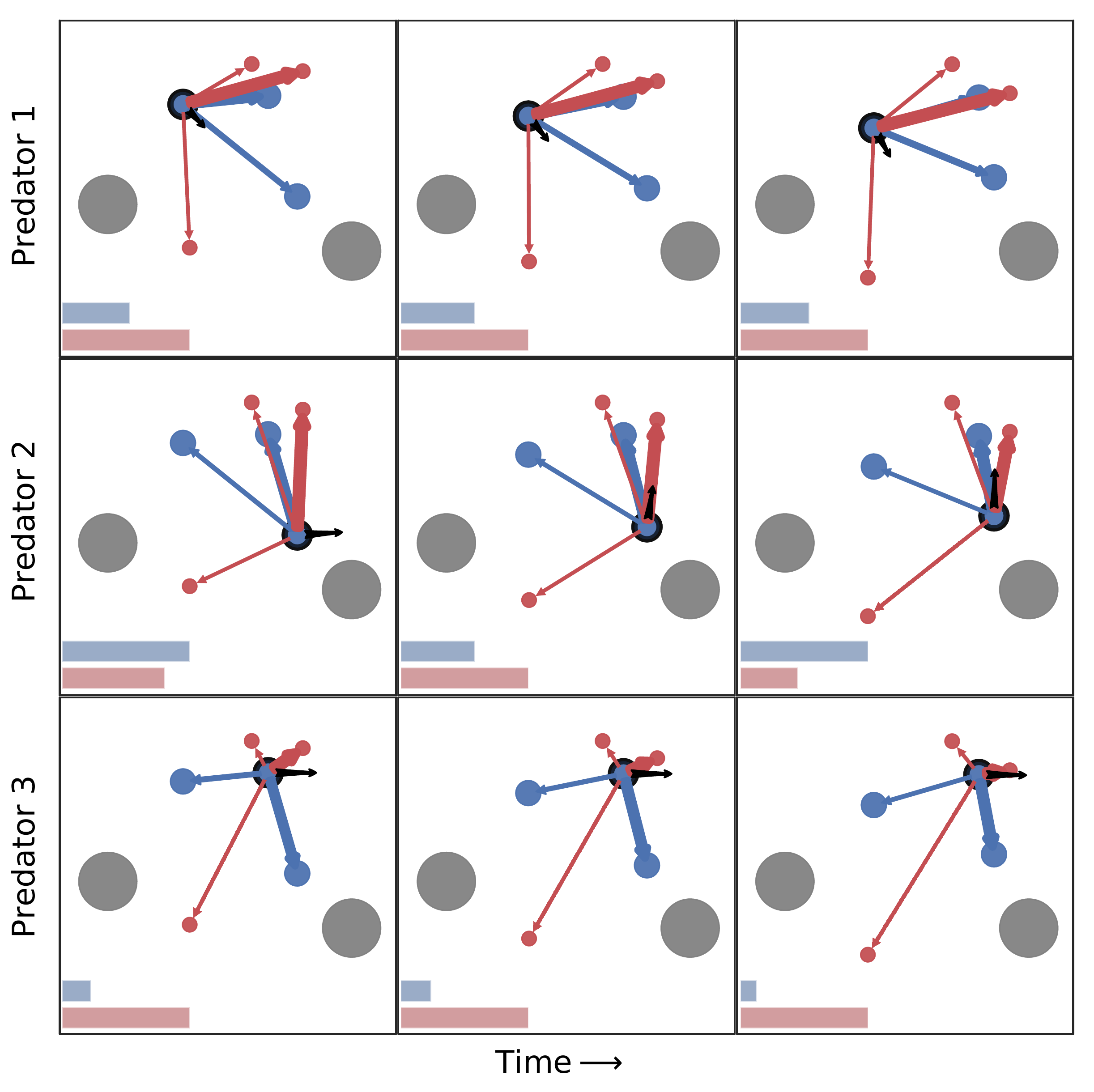}}
    \hspace{.11\textwidth}
    \subfigure[]{\label{fig:o}\includegraphics[width=46mm]{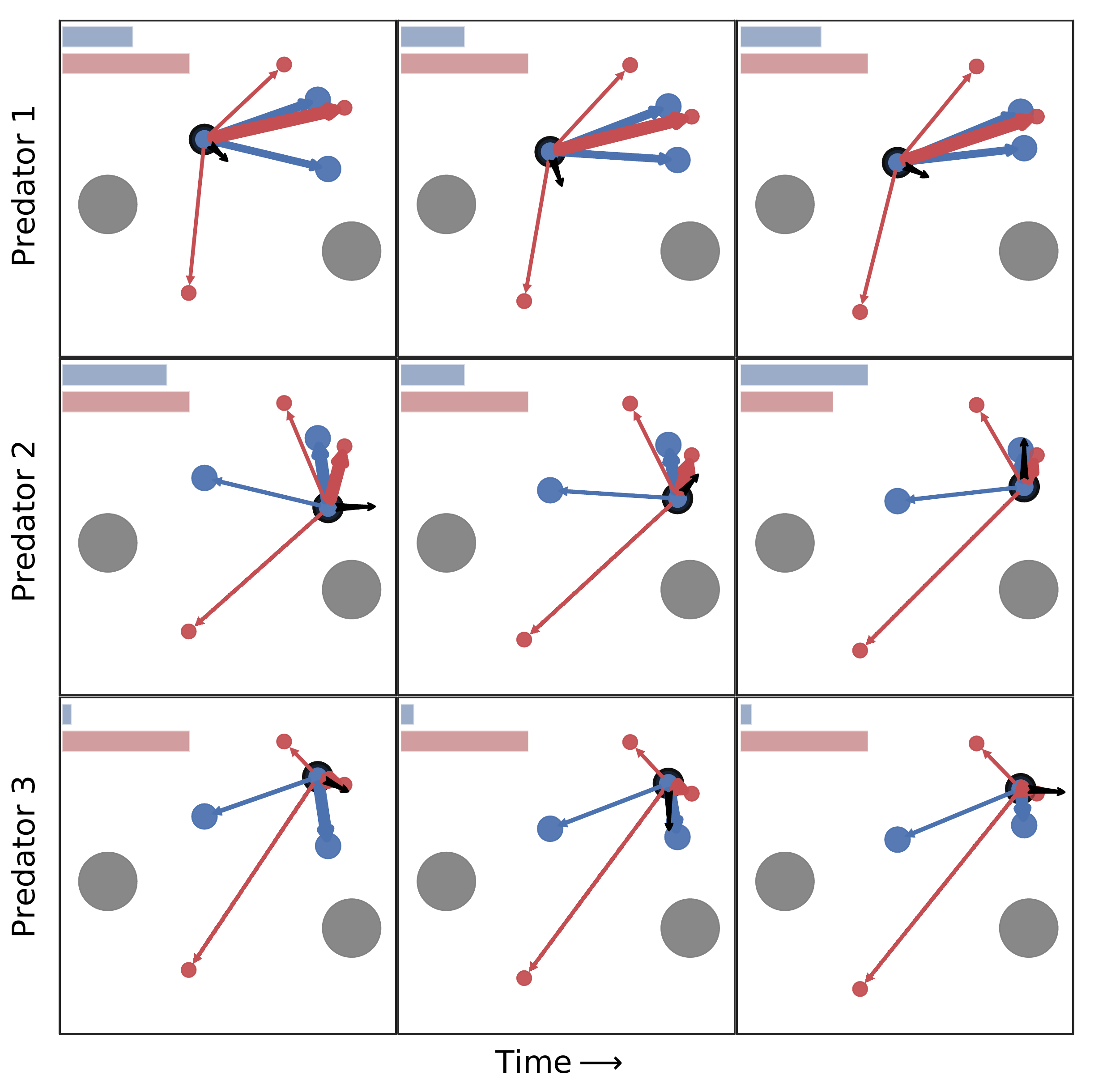}}
\end{figure*}

\section*{Supplementary Material}

% \subsection*{Interpreting Strategies for 3 vs. 3 predator-prey}
% We present the inter-agent and inter-group attentions of HAMA in another episode involving more steps in Figure 6. In (c), (j), and (o) in the figure, the preys are captured by the predators. When the chased prey is captured by the predators, the predators immediately move its attention to other preys that have not been captured yet. The already caught preys interrupt the predators, thus delaying the capture of other preys. In addition, the figures for the agent's attention over time can be converted into a video format for presentation.

% 
\subsection*{Multi-Agent Actor-Critic with Hierarchical Graph Attention Network Algorithm}
For completeness, we provide the HAMA algorithm below.

\begin{algorithm}[h!]
\caption{Multi-Agent Actor-Critic with HGAT Algorithm for $N$ agents}
\begin{algorithmic}[1]
\State Initialize actor networks $\mu$, critic networks $Q$, target networks $\mu'$ and $Q'$, replay buffer $\mathcal{D}$
\For {episode $= 1$ to $M$}
\State Initialize a random process $\mathcal{N}$ for action exploration
\State Receive initial observation $\mathbf{o}$
\For {$t = 1$ to $T$}
\State for each agent $i$ in agent group $k$, select action 
\State $a_i = \mu_k(o_i;\theta_k)+\mathcal{N}_t$
% \begin{align*}\qquad \qquad
% $a_i = \mu_k(o_i;\theta_k)+\mathcal{N}_t$\end{align*}
\State Execute actions $\mathbf{a}$ and receive reward 
\State $\mathbf{r}=(r_1,\dots,r_N)$ and new observation $\mathbf{o}'$
\State Store $(\mathbf{o},\mathbf{a},\mathbf{r},\mathbf{o}')$ in replay buffer $\mathcal{D}$
\State $\mathbf{o} \leftarrow \mathbf{o}'$
\For {$k = 1$ to $K$ for agent group $C^k$}
\For {agent $i = 1$ to $|C^k|$ in group $C^k$}
\State Sample a minibatch of $\mathcal{B}$ samples 
\State $(\mathbf{o}^j,\mathbf{a}^j,\mathbf{r}^j,\mathbf{o}'^j)$ from $\mathcal{D}$
\State \small Set $y^j_i=r^j_i+{\gamma}Q_k^{\mu'}({\mathbf{o}'^j},\mathbf{a}';\phi'_k)\arrowvert_{a'=\mu'(o'^j;\theta')}$
\State Update critic by minimizing the loss 
\State $\mathcal{L}(\phi_k)=\frac{1}{\mathcal{B}} \sum_j (Q_k^{\mu}({\mathbf{o}^j},\mathbf{a}^j;\phi_k)-y^j_i)^2$
\State Update actor using the sampled policy \State gradient:
\State $\nabla_{\theta_k}\mathcal{J}(\theta_k)
\approx$ \\ $\frac{1}{\mathcal{B}} \sum_j  \nabla_{\theta_k}\mu_k(o^j_i;\theta_k)\nabla_{a_i}Q_k^{\mu}({\mathbf{o}^j},\mathbf{a}^j;\phi_k )\arrowvert_{a_i=\mu_k(o^j_i;\theta_k)}$
\EndFor
\EndFor
\State Update target network parameters for each agent 
\State group $C^k$:
\begin{align*}
\phi_k' \leftarrow \tau\phi_k+(1-\tau)\phi_k'\\
\theta_k' \leftarrow \tau\theta_k+(1-\tau)\theta_k'
\end{align*}
\EndFor
\EndFor
\end{algorithmic}
\end{algorithm}

\subsection*{Hyperparameters for Experiments}
For the experiments in the environments, all the functions including embedding and attention functions in the actor and critic networks comprise a two-layered MLP with 256 units and a ReLU for nonlinear activation. The learning rate of Adam optimizer for updating the actor and critic network parameters is $1e-4$ and $1e-3$, respectively. The discount factor is 0.95. The target networks are updated with $\tau = 1e-3$ as a soft target. The network parameters that are initialized from Xavier uniform initialization are updated with a mini-batch of 1024. The size of experience replay buffer is $10^6$, and the parameters of the networks are updated every 100 samples added to the buffer after 2.5\% of the replay buffer is filled. The output layer of actor network provides the action as a five-sized tensor for hold, right, left, up, and down. We used three different random seeds and 120, 000 episodes with 25 steps (total 3 million steps) for training the networks in the environments. Subsequently, the trained models are validated in 200 episodes. 
% The experiments in this study were conducted based on the environments
% provided by \url{https://github.com/openai/multiagent-particle-envs}. All codes used in
% the experiments will be released.

\begin{table}[t]
  \caption{The mean and standard deviation of scores for predators in the more-the stronger game.}
  \label{table:table5}
  \centering
    \begin{tabular}{cccc}
        \toprule
    
        \multicolumn{1}{c}{predator} & \multicolumn{3}{c}{prey $(n=3)$}\\
        \cmidrule(r){2-4}
        
        \multicolumn{1}{c}{$(n=3)$} & \multicolumn{1}{c}{MADDPG} & \multicolumn{1}{c}{MAAC} & \multicolumn{1}{c}{HAMA}\\
        \midrule
        % \multirow{5}{4em}{predator \\$(n=3)$} &
        \multicolumn{1}{c}{Heuristic1} & 0.19\scriptsize$\pm$0.05 & 0.17\scriptsize$\pm$0.09 & 0.003\scriptsize$\pm$0.001\\
        \multicolumn{1}{c}{Heuristic2}  & 0.73\scriptsize$\pm$0.10 & 0.49\scriptsize$\pm$0.18&0.005\scriptsize$\pm$0.001\\
        \multicolumn{1}{c}{MADDPG}  & 1.64\scriptsize$\pm$0.13 & 1.81\scriptsize$\pm$0.29 &0.02\scriptsize$\pm$0.001\\      \multicolumn{1}{c}{MAAC}   & 0.77\scriptsize$\pm$0.22 & 0.71\scriptsize$\pm$0.23 & 0.64\scriptsize$\pm$0.20\\
        \multicolumn{1}{c}{HAMA}    & \textbf{5.45}\scriptsize$\pm$0.12 & \textbf{3.45}\scriptsize$\pm$0.30 &\textbf{2.08}\scriptsize$\pm$0.13\\
    
        \bottomrule
    \end{tabular}
\end{table}

\subsection*{Results for The More-The Stronger}

The more-the stronger game keeps the framework of the 3 vs. 3 predator-prey game. The additional game rule in this game is that when the preys are clustered together, only a group of predators whose size is equal to or larger than that of the clustered preys can capture the preys. For example, one predator can capture one prey by itself, but three-gathered predators are required to capture three-gathered preys. The more predators are gathered, the stronger is the power; the more preys are gathered, the stronger is the shield. When fewer predators approach the gathered preys, the predators cannot capture the preys. Therefore, the agents in the game require more complex strategies to achieve their objectives. The rewards of the predators and preys in the game are the same as those in the 3 vs. 3 predator-prey game. The results are summarized in Table \ref{table:table5}. As shown in the table, HAMA outperforms other models in this game as well. 

\subsection*{Interpreting Strategies for 3 vs. 3 Predator-Prey}
We present the inter-agent and inter-group attentions of HAMA in another episode involving more steps in Figure 6. In (c), (j), and (o) in the figure, the preys are captured by the predators. When the chased prey is captured by the predators, the predators immediately move their attention to other preys that have not been captured yet. The already caught preys interrupt the predators, thus delaying the capture of other preys. 
% In addition, the figures for the agent's attention over time can be converted into a video format for presentation.

\subsection*{Learning Curves on 3 vs. 3 Predator-Prey}
\begin{figure}
  \centering
  \includegraphics[width=0.4\textwidth]{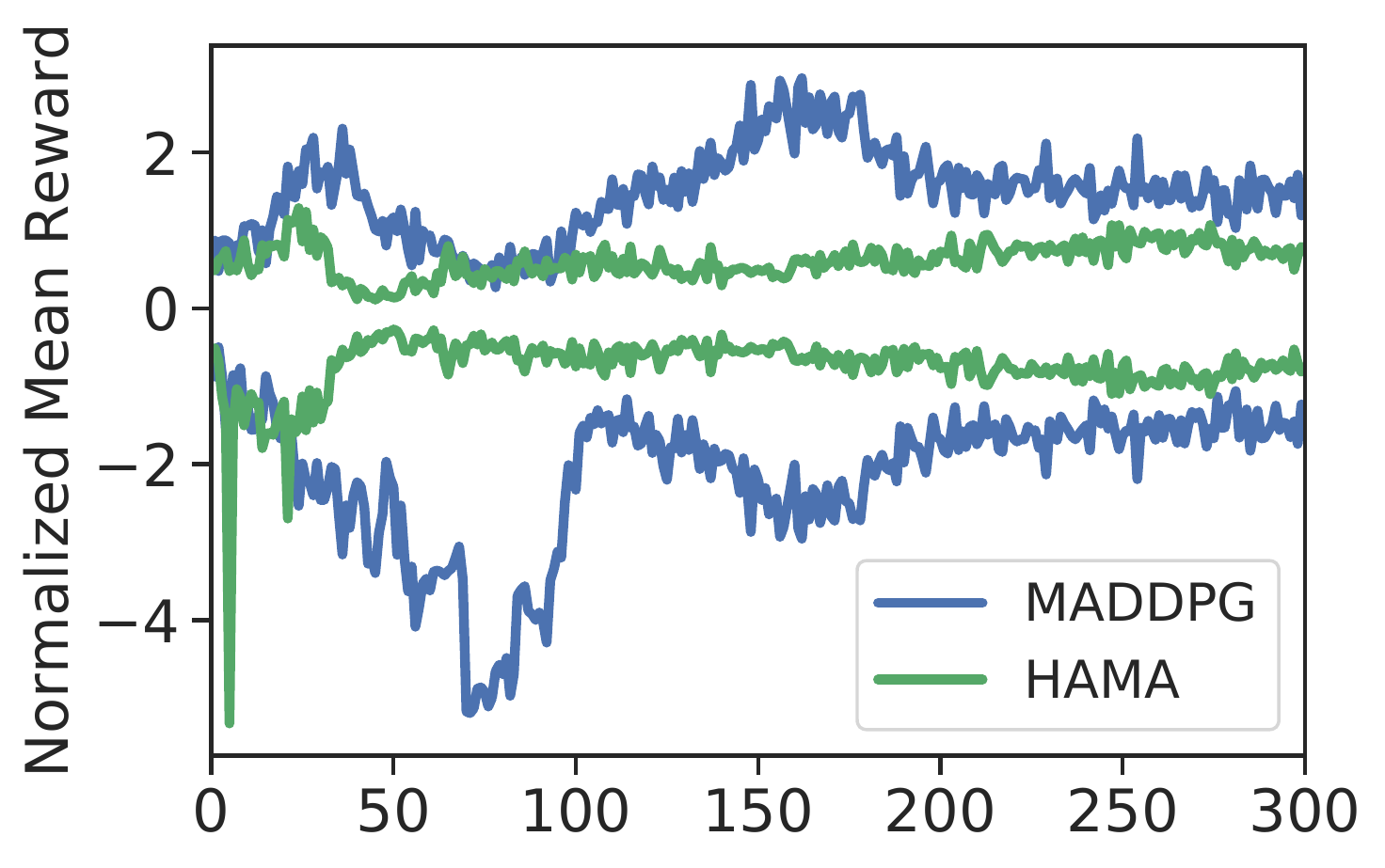}
  \caption{Rewards during training on 3 vs. 3 predator-prey.}
  \label{fig:fig6}
\end{figure} 
In Figure \ref{fig:fig6}, the rewards over 3 million steps are averaged every 10,000 steps and normalized. The positive and negative rewards of each model represent the rewards of the predators and preys, respectively. The rewards of the predators and preys are symmetric but not completely symmetric owing to the penalty obtained by the preys when they are outside of a particular zone. In the figure, the predators of MADDPG exhibit a higher score than the predators of HAMA during training. When HAMA is considered to outperform MADDPG from the primary results in the main paper, we assume that the more intelligent preys have learned to counter the intelligent predators in HAMA. Furthermore, in the early stages of training, the preys leave a certain zone to escape from the predators, which appears to be extremely negative in the early rewards for the preys. HAMA exhibits this phenomenon much earlier than MADDPG. We assume that this is because HAMA trains the given samples more efficiently than MADDPG.

\end{document}